\documentclass{article}

\usepackage[preprint]{neurips_2026}

\usepackage[utf8]{inputenc} 
\usepackage[T1]{fontenc}    
\usepackage{hyperref}       
\usepackage{url}            
\usepackage{booktabs}       
\usepackage{amsfonts}       
\usepackage{nicefrac}       
\usepackage{microtype}      
\usepackage{xcolor}         
\usepackage{amsmath}
\usepackage{amssymb}
\usepackage{amsthm}
\usepackage{thmtools, thm-restate}
\usepackage{mathtools}
\usepackage{stmaryrd}
\usepackage{dsfont}
\usepackage{enumitem}
\usepackage{listings}
\usepackage{subcaption}
\usepackage[ruled,vlined]{algorithm2e}
\usepackage{tikz}
\usetikzlibrary{arrows.meta}

\newtheorem{theorem}{Theorem}
\newtheorem{lemma}[theorem]{Lemma}

\newtheorem{defn}[theorem]{Definition}

\newtheorem{prop}[theorem]{Proposition}

\lstdefinelanguage{pseudocode}{
  morekeywords={for, if, then, input, while, do, else, return}
}
\lstdefinestyle{pseudocode}{
  language=pseudocode,
  basicstyle=\small\ttfamily,
  keywordstyle=\bfseries,
  numbers=left,
  numberstyle=\tiny,
  numbersep=5pt,
  columns=fullflexible,
  breaklines=true
}

\newcommand{\bN}{\mathbf{N}}
\newcommand{\bX}{\mathbf{X}}
\newcommand{\bv}{\mathbf{v}}
\def\cA{{\mathcal A}}

\def\cE{{\mathcal E}}
\def\cF{{\mathcal F}}
\def\cG{{\mathcal G}}
\def\cH{{\mathcal H}}
\def\cI{{\mathcal I}}
\def\cL{{\mathcal L}}

\def\cP{{\mathcal P}}

\def\cS{{\mathcal S}}

\def\R{{\mathbb R}}
\newcommand{\A}{\mathcal{A}}

\newcommand{\C}{\mathcal{C}}
\newcommand{\D}{\mathcal{D}}
\newcommand{\E}{\mathbb{E}}
\newcommand{\Chat}{\widehat{\mathcal{C}}}
\newcommand{\Ehat}{\widehat{\mathcal{E}}}
\newcommand{\Phat}{\widehat{\mathcal{P}}}
\newcommand{\Ihat}{\widehat{\mathcal{I}}}
\newcommand{\Ghat}{\widehat{G}}

\newcommand{\one}{\mathds{1}}

\DeclareMathOperator{\Var}{Var}

\DeclareMathOperator{\diag}{diag}

\DeclareMathOperator{\dist}{dist}
\DeclareMathOperator{\FD}{FD}
\DeclareMathOperator{\FDR}{FDR}
\DeclareMathOperator{\FNR}{FNR}
\DeclareMathOperator{\FPP}{FPP}
\DeclareMathOperator{\Ad}{Ad}

\DeclareMathOperator{\Pa}{Pa}
\DeclareMathOperator{\Ne}{Ne}
\DeclareMathOperator{\Desc}{Desc}

\title{Causal Discovery with False Positive Error Control}

\author{%
  Erik Jahn \\
  Computing and Mathematical Sciences\\
  California Institute of Technology\\
  Pasadena, CA 91125 \\
  \texttt{ejahn@caltech.edu} \\
  \And
  Venkat Chandrasekaran \\
  Computing and Mathematical Sciences\\
  California Institute of Technology\\
  Pasadena, CA 91125 \\
  \texttt{venkatc@caltech.edu} \\
  \And 
  Frederick Eberhardt \\
  Humanities and Social Sciences\\
  California Institute of Technology\\
  Pasadena, CA 91125 \\
  \texttt{fde@caltech.edu} \\
  \And
  Leonard Schulman \\
  Computing and Mathematical Sciences\\
  California Institute of Technology\\
  Pasadena, CA 91125 \\
  \texttt{schulman@caltech.edu} \\
}

\begin{document}

\maketitle

\begin{abstract}
  Causal discovery methods produce data-driven hypotheses about causal relationships, which may become new scientific findings or inform decisions about downstream interventions. In such settings, false positive causal claims can be more consequential than false negatives. Yet few methods target false positive error control in causal discovery, and existing error guarantees are often asymptotic and require strong versions of the faithfulness assumption. We introduce the $m$-conservative PC algorithm and prove that it controls false discovery rate in the linear Gaussian setting under an extremely weak version of the faithfulness assumption. In our framework, false positive error for causal graphs is defined based on the conditional-independence information they entail, yielding a metric that penalizes both false adjacencies and incorrect edge orientations. The parameter $m$ quantifies a trade-off between algorithmic complexity and the strength of the required faithfulness assumption. The $m$-conservative PC algorithm combines undirected graph learning, a Benjamini-Yekutieli-style procedure for conditional independence testing, and a conservative orientation step that marks unresolved or contradictory edge directions as ambiguous.  Our algorithm produces valid outputs and retains error guarantees even in the finite-sample setting, despite possible random errors in the conditional independence tests.
\end{abstract}

\section{Introduction}

Causal discovery algorithms aim to infer causal relationships from observational data and are being applied across many scientific fields, such as biology, astrophysics, and economics~\citep{kelly_review_2022, jin_causal_2025, addo_exploring_2021}. 
Novel scientific findings often claim the presence of a new causal effect, so false positive error control becomes an important objective. 

The existing approaches to false positive error control learn causal graphs using conditional independence tests~\citep{li_controlling_2009, strobl_estimating_2019} and are derived from the classic PC algorithm~\citep{spirtes_causation_2001}. Both approaches target the false discovery rate rather than the absolute number of errors, which often allows for higher statistical power~\citep{benjamini_controlling_1995}. However, the typical notion of false positive error for causal graphs only focuses on the existence of claimed causal connections and does not effectively score errors in their orientations. In this paper, we adopt a recently proposed framework~\citep{TBC24} that addresses this issue and gives a unifying definition for false positive error across many different model selection problems. For causal graphs this approach counts any edge as a false positive that corresponds to a false conditional dependence claim entailed by the discovered causal graph. The resulting measure naturally accounts for both the existence and orientation of false edges, without penalizing orientations that are not identifiable from observational data. 

While the PC algorithm can be shown to asymptotically recover the true causal graph up to its Markov equivalence class, any finite sample error guarantee relies on the strong faithfulness assumption~\citep{kalisch_estimating_2007}. As this assumption can often be violated in practice~\citep{uhler_geometry_2013}, developing causal discovery algorithms with error guarantees under weaker assumptions is an ongoing research problem~\citep{spirtes_uniformly_2014, zhalama_sat-based_2017, ng_reliable_2021}. Under such weaker assumptions, the Markov equivalence class of the true causal graph may no longer be identifiable, so some algorithms output e-patterns, which are graphical representations of sets of directed acyclic graphs that can span multiple Markov equivalence classes~\citep{ramsey_adjacency-faithfulness_2006, lemeire_conservative_2012}. Moreover, it has been shown that faithfulness violations can become undetectable when so-called triangle faithfulness is violated, suggesting a fundamental limitation for absolute error guarantees~\citep{zhang_detection_2008}. However, under the weaker objective of false positive error control we show that error guarantees for causal discovery are possible even for versions of faithfulness weaker than triangle faithfulness.  

\textbf{Contributions.}
Consider a linear Gaussian causal model that induces the probability distribution $P^*$ and whose nonzero direct causal effects are represented by a directed acyclic graph (DAG) $G^*$. We introduce \emph{$m$-connected-collider faithfulness}, which gives a new family of increasingly weaker faithfulness conditions, parametrized by $m$. We prove that $m$-connected-collider faithfulness can be directly linked to controlling false negative edges in the undirected graph $H^*$ corresponding to $P^*$. Moreover, we introduce the $m$-conservative PC algorithm that learns an e-pattern representing a set of DAGs, and we prove that the expected minimal false discovery rate across these DAGs is controlled under the following assumptions:

\begin{enumerate}[label=(\arabic*)]
    \item the true causal model satisfies $m$-connected-collider faithfulness; \label{ass:mccf}
    \item every node in $G^*$ is adjacent to at most $\Delta$ other nodes;\label{ass:sparsity}
    \item we have access to an undirected graph learning method that fails to produce a supergraph of $H^*$ with probability at most $p$; \label{ass:uglearning}
    \item we have access to $p$-values for conditional independence testing (that are valid independently of the failure event in assumption~\ref{ass:uglearning}). \label{ass:p-values}
\end{enumerate}

\begin{restatable}{theorem}{mainthm}\label{thm:main}
    Suppose Assumptions~\ref{ass:mccf} to \ref{ass:p-values} are satisfied. Then, the output $\Ehat$ of the $m$-conservative PC algorithm with significance level $\alpha$ satisfies $\E\left[\FDR(\Ehat, [G^*])\right] \leq (\Delta+1) \cdot \alpha + p$.
\end{restatable}

We also demonstrate the empirical performance of the $m$-conservative PC algorithm with synthetic examples and data from the Causal Chamber light tunnel~\citep{gamella_causal_2025}. 

\section{Background}
\subsection{Linear Gaussian structural equation models}\label{sec:SEMs}
A linear Gaussian structural equation model (SEM) for a vector $\bX = (X_1, \dots, X_n)$ is given by a strictly lower-triangular matrix 
$A \in \R^{n \times n}$ and a positive definite noise covariance matrix $\Sigma_\bN$ that together define the structural equations
\begin{align*}
    \bX = A \bX + \bN,
\end{align*}
where $\bN \sim \mathcal{N}(0, \Sigma_\bN)$. Since $A$ is strictly lower-triangular, $(I-A)$ is invertible and we obtain $\bX = (I-A)^{-1} \bN$.
In particular, $\bX$ is Gaussian with mean $0$ and covariance matrix
\begin{align}
    \Sigma_\bX = (I - A)^{-1} \Sigma_\bN (I-A)^{-\top}. \label{eq:sigmas}
\end{align}

Throughout the paper, we assume causal sufficiency, which means that the entries of the noise vector $\bN$ are mutually independent.
Hence, we have $\Sigma_\bN = \diag(\bv)$ where $\bv = (v_1, \dots, v_n)$ is a vector of variances $v_i = \Var(N_i) > 0$. In the following, denote $[n] = \{1, \dots, n\}$.

\begin{defn} \label{defn:SEM-graphs}
    Let $(A, \bv)$ be a linear Gaussian SEM. 
    \begin{enumerate}
        \item We define the corresponding DAG $G = G(A)$ to be the DAG on vertex set $[n]$ with an edge $i \to j$ if and only if $i < j$ and $A_{ji} \neq 0$.
        \item We define the corresponding undirected graph $H = H(A, \bv)$ to be the graph on vertex set $[n]$ with an edge $\{i,j\}$ if and only if $(\Sigma_\bX^{-1})_{ij} \neq 0$.
    \end{enumerate}
    \end{defn}

\subsection{Graphical models and false positive error} \label{sec:FPE-background}
We use standard terminology for d-separation, Markov-equivalence and CPDAGs (see Appendix~\ref{apx:background} for details). When evaluating causal discovery algorithms, it is unclear how incorrect edge orientations should count towards false positive error. Recently, \citet{TBC24} introduced a framework for defining false positive error that applies across many different model selection problems, including those that lack a clear binary structure, only requiring a partial ordering on the set of models. Causal discovery algorithms typically aim to recover the CPDAG that represents the Markov equivalence class of the true causal graph, accounting for the fact that identification of the causal ground truth beyond Markov equivalence is often impossible without additional assumptions. Since Markov equivalence classes are defined in terms of d-separations, a natural partial ordering on the set of CPDAGs is in terms of inclusion of the lists of d-separation relationships that they represent. 
\begin{defn}[partial ordering on CPDAGs] \label{defn:cpdag-poset}
    Let $\mathcal{C}, \mathcal{D}$ be CPDAGs on a common vertex set $V$. We define $\C \preceq  \D$ if for all $v,w \in V$ and $Z \subseteq V \setminus \{v,w\}$, we have $(v \perp w \mid Z)_{\D} \implies (v \perp w \mid Z)_{\C}$.
\end{defn}
Now, the key idea is to say that $\widehat{G}$ has no false positive error with respect to $G^*$ if and only if $[\widehat{G}] \preceq [G^*]$, where $[G]$ denotes the CPDAG representing the Markov equivalence class of $G$. In other words, $\widehat{G}$ does not entail any conditional dependencies that are not already entailed by $G^*$. For instance, this implies that $\widehat{G}$ must neither have false positive adjacencies nor false unshielded colliders (that are unshielded non-colliders in $G^*$). Hence, this notion naturally captures both the misdirection and the false inclusion of edges. Now, a natural quantification of false positive error is given as follows:

\begin{defn}[false positive error~\citep{TBC24}] \label{defn:FD}
    For two CPDAGs $\widehat{\C}$ and $\C^*$, we define the false discovery $\FD$ and the false discovery rate $\FDR$ as 
    \begin{align*}
    \FD(\widehat{\C}, \C^*) := \Ad(\widehat{\C})- \max_{\A \preceq \widehat{\C}, \A \preceq \C^*}\Ad(\A)  , \qquad \FDR(\widehat{\C}, \C^*) := \frac{\FD(\widehat{\C}, \C^*)}{\max(\Ad(\widehat{\C}), 1)}.        
    \end{align*}
\end{defn}
Here, $\Ad(\C)$ denotes the number of adjacencies, that is directed and undirected edges, of $\C$\footnote{One can show that $\Ad(\C)$ is a valid rank function for the poset of CPDAGs ordered by $\preceq$~\citep{Chickering03}, which motivates using rank differences as our notion of distance here (see also Appendix~\ref{apx:background}).}. Intuitively, the false discovery of $\widehat{\C}$ with respect to $\C^*$ is the minimal number of adjacencies of $\widehat{\C}$ that need to be deleted to arrive at a CPDAG that has no false discovery anymore with respect to $\C^*$. The following observation is crucial for using these definitions in practice: even though the list of d-separation statements entailed by a CPDAG may have exponential size, one can check if $\C \preceq \D$ in polynomial time with the characterization below:

\begin{lemma}[\citep{kocka_characterizing_2001}]\label{lem:char_inclusion}
    Let $\mathcal{C}, \mathcal{D}$ be CPDAGs on $[n]$ and let $G$ be a consistent DAG-extension of $\D$ with lower triangular adjacency matrix. We have that $\C \preceq \D$ if and only if $(i \perp j \mid \Pa(j, G))_{\C}$ for all pairs $i < j\in [n]$ such that $i \notin \Pa(j, G)$.
\end{lemma}

\begin{figure}
    \centering
\begin{tikzpicture}[
    >=Stealth,
    scale=0.6,
    every node/.style={transform shape},
    dir/.style={->, thick, shorten >=1.5pt, shorten <=1.5pt},
    undir/.style={-, thick},
    amb/.style={blue,very thick}
]

\begin{scope}[shift={(0,0)}, x=1.35cm, y=1.35cm]

    \draw[amb] (-1,0) ++(-45:0.7cm)
    arc[start angle=-45, end angle=45, radius=0.7cm];
    \draw[amb] (1,0) ++(135:0.7cm)
    arc[start angle=135, end angle=225, radius=0.7cm];

    \node[circle, inner sep = 1pt, font = \Large] (1)  at (-1,0) {$X_1$};
    \node[circle, inner sep = 1pt, font = \Large] (2) at (0,1) {$X_2$};
    \node[circle, inner sep = 1pt, font = \Large] (3) at (1,0) {$X_3$};
    \node[circle, inner sep = 1pt, font = \Large] (4) at (0,-1) {$X_4$};

    \draw[undir] (1) -- (2);
    \draw[undir] (1) -- (4);
    \draw[undir] (3) -- (2);
    \draw[undir] (3) -- (4);
\end{scope}

\begin{scope}[shift={(4.2,1.0)}, x=0.75cm, y=0.75cm]
    \coordinate (L) at (-1,0);
    \coordinate (T) at (0,1);
    \coordinate (R) at (1,0);
    \coordinate (B) at (0,-1);

    \draw[dir] (T) -- (L);
    \draw[dir] (T) -- (R);
    \draw[dir] (R) -- (B);
    \draw[dir] (B) -- (L);
\end{scope}

\begin{scope}[shift={(6.0,1.0)}, x=0.75cm, y=0.75cm]
    \coordinate (L) at (-1,0);
    \coordinate (T) at (0,1);
    \coordinate (R) at (1,0);
    \coordinate (B) at (0,-1);

    \draw[dir] (T) -- (L);
    \draw[dir] (R) -- (T);
    \draw[dir] (B) -- (R);
    \draw[dir] (B) -- (L);
\end{scope}

\begin{scope}[shift={(7.8,1.0)}, x=0.75cm, y=0.75cm]
    \coordinate (L) at (-1,0);
    \coordinate (T) at (0,1);
    \coordinate (R) at (1,0);
    \coordinate (B) at (0,-1);

    \draw[dir] (T) -- (L);
    \draw[dir] (R) -- (T);
    \draw[dir] (R) -- (B);
    \draw[dir] (B) -- (L);
\end{scope}

\begin{scope}[shift={(10,0.0)}, x=0.75cm, y=0.75cm]
    \coordinate (L) at (-1,0);
    \coordinate (T) at (0,1);
    \coordinate (R) at (1,0);
    \coordinate (B) at (0,-1);

    \draw[dir] (T) -- (L);
    \draw[dir] (T) -- (R);
    \draw[dir] (B) -- (R);
    \draw[dir] (B) -- (L);
\end{scope}

\begin{scope}[shift={(4.2,-1.0)}, x=0.75cm, y=0.75cm]
    \coordinate (L) at (-1,0);
    \coordinate (T) at (0,1);
    \coordinate (R) at (1,0);
    \coordinate (B) at (0,-1);

    \draw[dir] (T) -- (L);
    \draw[dir] (T) -- (R);
    \draw[dir] (B) -- (R);
    \draw[dir] (L) -- (B);
\end{scope}

\begin{scope}[shift={(6.0,-1.0)}, x=0.75cm, y=0.75cm]
    \coordinate (L) at (-1,0);
    \coordinate (T) at (0,1);
    \coordinate (R) at (1,0);
    \coordinate (B) at (0,-1);

    \draw[dir] (L) -- (T);
    \draw[dir] (T) -- (R);
    \draw[dir] (B) -- (R);
    \draw[dir] (B) -- (L);
\end{scope}

\begin{scope}[shift={(7.8,-1.0)}, x=0.75cm, y=0.75cm]
    \coordinate (L) at (-1,0);
    \coordinate (T) at (0,1);
    \coordinate (R) at (1,0);
    \coordinate (B) at (0,-1);

    \draw[dir] (L) -- (T);
    \draw[dir] (T) -- (R);
    \draw[dir] (B) -- (R);
    \draw[dir] (L) -- (B);
\end{scope}

\end{tikzpicture}
    \caption{The ambiguous unshielded triples in the e-pattern to the left are marked with blue arcs. To the right are the $7$ DAGs represented by the e-pattern.}
    \label{fig:example0}
\end{figure}
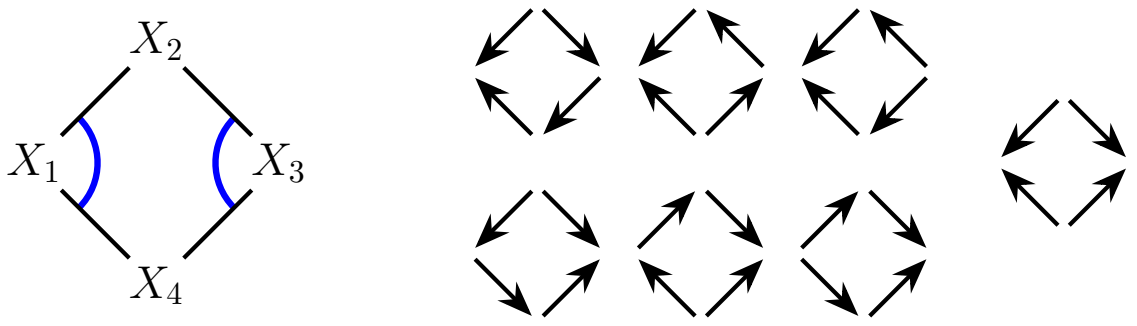

We will see later that faithfulness violations can make even the CPDAG of the true causal graph unidentifiable from observational samples. Therefore, we will present an algorithm that typically outputs \emph{e-patterns}, which can represent DAGs from multiple Markov equivalence classes at once. According to \citet{ramsey_adjacency-faithfulness_2006}, we define an e-pattern as a partially directed graph, for which some unshielded triples are marked as ambiguous. Edges that belong to ambiguous triples are always undirected. An e-pattern $\mathcal{E}$ represents all DAGs $G$ that have the same adjacencies as $\cE$, contain all directed edges of $\mathcal{E}$, and satisfy the condition that every unshielded collider in $G$ is either directed as a collider in $\mathcal{E}$ or the triple is marked as ambiguous. For an e-pattern $\cE$ and a CPDAG $\C^*$, we define the false discovery to be $\FD(\cE, \C^*) = \min_{G} \FD([G], \C^*)$ where $G$ runs over all DAGs represented by $\cE$. Similarly, the false discovery rate is defined as $\FDR(\cE, \C^*) = \FD(\cE, \C^*) / \max(\Ad(\cE), 1)$, where $\Ad(\cE)$ denotes the number of adjacencies of $\cE$. We give an example of an e-pattern and its represented DAGs in Figure~\ref{fig:example0}.

\section{Connected-collider faithfulness} \label{sec:faithfulness}
In this section, we introduce the family of weak faithfulness conditions, under which our algorithm provides false positive error control. For a DAG $G$ on $[n]$, and a pair of vertices $\{i,j\}$ let us denote the \emph{set of collider vertices based at $\{i,j\}$} as $C(i,j) := \{k \in [n] \mid i \to k, j \to k \}$. 

\begin{defn} \label{defn:faithful_colliders}
    Let $(A, \bv)$ be a linear Gaussian SEM and consider
    an edge $i \to j$ in $G(A)$. We say that the set of colliders based at $(i,j)$ is faithful if $\sum_{k \in C(i,j)} A_{ki} A_{kj} v_k^{-1} \neq A_{ji}v_j^{-1}$.
\end{defn}

The significance of this condition stems from the fact that it controls the adjacencies in the undirected graph of the SEM. 

\begin{restatable}{lemma}{faithfultriangles}\label{lem:faithful_triangles}
    Let $(A, \bv)$ be a linear Gaussian SEM and let $i \to j$ be an edge in $G(A)$. 
    Then, $\{i,j\}$ is an edge in $H(A, \bv)$ if and only if the set of colliders based 
    at $(i,j)$ is faithful. 
\end{restatable}

The proof of this equivalence is straightforward (see Appendix~\ref{apx:proofs}). We are now ready to introduce a new family of faithfulness conditions parametrized by a positive integer $m$. 

\begin{defn} \label{defn:m-faithfulness}
We say that a linear Gaussian SEM $(A, \bv)$ satisfies the \emph{$m$-connected-collider faithfulness} condition if any subgraph of $G(A)$ with a unique root that has at least $m$ edges contains an edge $i \to j$ such that the set of colliders based at $(i,j)$ is faithful. 
\end{defn}

Note that the $m$-connected-collider faithfulness condition becomes strictly weaker as $m$ increases and eventually vacuous (for $m > \binom{n}{2}$). Moreover, for $m=1$, it is already weaker than \emph{triangle faithfulness} (see Appendix~\ref{apx:background}), which follows from the facts that the set of colliders based at an edge $\{i,j\}$ is non-empty only if there is a triangle at $\{i,j\}$ and it is faithful if and only if $X_i \not \perp X_j \mid X_{V \setminus \{i,j\}}$. See Figure~\ref{fig:faithfulness} for an empirical study of the frequency of near-violations for the different faithfulness notions. The $m$-connected-collider faithfulness condition is motivated by the following property of the undirected graph corresponding to the SEM:

\begin{restatable}{prop}{distance}\label{prop:distance-m}
    Let $(A, \bv)$ be a linear Gaussian SEM, satisfying $m$-connected-collider faithfulness for $m \geq 1$. If $i,j$ are adjacent in $G(A)$, then they have distance at most $m$ in $H(A, \bv)$.
\end{restatable}

 The concept of exploiting a condition on the undirected graph for causal discovery is not new, and indeed the $m$-connected-collider faithfulness condition with $m=1$ appears as Assumption 1 in~\citet{loh_high-dimensional_2014} and as the single shielded-collider-faithfulness condition in~\citet{ng_reliable_2021}. In fact, Proposition~\ref{prop:distance-m} can be seen as a natural generalization of Theorem 2 in~\citet{ng_reliable_2021}. We state the complete proof in Appendix~\ref{apx:proofs}.

\section{The $m$-conservative PC algorithm} \label{sec:mcpc}

\subsection{Selecting conditional independence tests}
\begin{figure}
    \centering
    \begin{tikzpicture}[scale=0.8,    >=Stealth,    node/.style={circle, inner sep=1pt, font=\large},    edge/.style={->, thick},  undirected/.style={-, thick},  lab/.style={font=\normalsize}, 
    amb/.style={very thick, blue}]

\node[node] (X1) at (0,0.75) {$X_1$};
\node[node] (X2) at (1.4,1.5) {$X_2$};
\node[node] (X3) at (1.4,0) {$X_3$};
\node[node] (X4) at (2.8,0.75) {$X_4$};

\draw[edge] (X1) -- node[above left, lab] {$A_{21}$} (X2);
\draw[edge] (X1) -- node[below left, lab] {$A_{31}$} (X3);
\draw[edge] (X2) -- node[right, lab] {$A_{32}$} (X3);
\draw[edge] (X2) -- node[above right, lab] {$A_{42}$} (X4);
\draw[edge] (X3) -- node[below right, lab] {$A_{43}$} (X4);

\node[node] at (1.4,-0.7) {$G^*$};

\node[node] (Y1) at (4.3,0.75) {$X_1$};
\node[node] (Y2) at (5.7,1.5) {$X_2$};
\node[node] (Y3) at (5.7,0) {$X_3$};
\node[node] (Y4) at (7.1,0.75) {$X_4$};

\draw[edge] (Y1) -- (Y2);
\draw[edge] (Y1) -- (Y3);
\draw[undirected] (Y2) -- (Y3);
\draw[edge] (Y4) -- (Y2);
\draw[edge] (Y4) -- (Y3);

\node[node] at (5.7,-0.7) {$\Chat$};

\node[node] (E1) at (8.6,0.75) {$X_1$};
\node[node] (E2) at (10.0,1.5) {$X_2$};
\node[node] (E3) at (10.0,0) {$X_3$};
\node[node] (E4) at (11.4,0.75) {$X_4$};

\draw[undirected] (E1) -- (E2);
\draw[undirected] (E1) -- (E3);
\draw[undirected] (E2) -- (E3);
\draw[undirected] (E2) -- (E4);
\draw[undirected] (E3) -- (E4);

    \draw[amb] (10,0) ++(30:0.6cm)
    arc[start angle=30, end angle=150, radius=0.6cm];
    \draw[amb] (10,1.5) ++(-30:0.6cm)
    arc[start angle=-30, end angle=-150, radius=0.6cm];

\node[node] at (10.0,-0.7) {$\Ehat$};

\end{tikzpicture}

    \caption{Suppose the linear Gaussian SEM $(A, \mathbf{1})$ with corresponding DAG $G^*$ satisfies $A_{21}(A_{42}+A_{32}A_{43}) + A_{31}A_{43} = 0$, implying $X_1 \perp X_4$. Then, the output of the PC algorithm given the exact distribution of the model is $\widehat{\C}$, which has $\FD(\widehat{\C}, [G^*]) = 2$ due to two incorrect unshielded colliders. The conservative PC and the $m$-conservative PC algorithm (for all $m$) output the e-pattern $\Ehat$ that represents the true DAG $G^*$.}
    \label{fig:example1}
\end{figure}
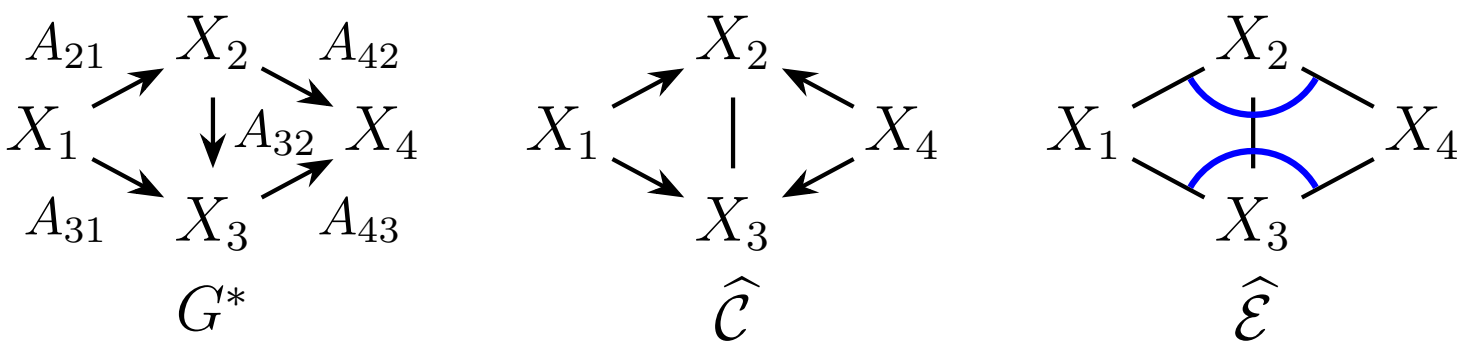

One of the main challenges that constraint-based causal discovery algorithms have to address is finding the most informative conditional independencies in the data without conducting all of the exponentially many possible conditional independence tests. For a DAG $G$ corresponding to a linear Gaussian SEM on $\bX$, we know that by the Markov property (Proposition~\ref{prop:CMC} in Appendix~\ref{apx:background}), each d-separation statement of a DAG can be found as a conditional independency in the probability distribution of $\bX$. Crucially, knowing exactly one true d-separation statement for each non-adjacent pair of vertices in $G$ is sufficient to recover the full CPDAG of $G$. Moreover, one such d-separation statement can always be found by conditioning on the parent set of the correct vertex (Lemma~\ref{lem:parentseparation}). This motivates two design choices of the PC algorithm (Algorithm~\ref{alg:pc} in Appendix~\ref{apx:background}):
\begin{enumerate}[label=(\arabic*)]
    \item stop performing conditional independence tests for a pair $X_i, X_j$ once a conditional independency for this pair was found; \label{design-choice-1}
    \item keep and update an estimated adjacency graph for $G$, and only test independence of $X_i, X_j$ conditioned on sets of current neighbors in the adjacency graph. \label{design-choice-2}
\end{enumerate}
While improving efficiency, both these choices can lead to failures under (near-)faithfulness violations. In the example in Figure~\ref{fig:example1}, the PC algorithm finds the conditional independency $X_1 \perp X_4$ first, which does not correspond to a correct d-separation, and therefore leads to incorrect collider assignments. The conservative PC algorithm~\citep{ramsey_adjacency-faithfulness_2006} (Algorithm~\ref{alg:cpc} in Appendix~\ref{apx:background}) addresses this potential failure by abandoning~\ref{design-choice-1} and checking for all possible separating sets among the neighbors of two vertices. If two separating sets lead to conflicting collider assignments, then the corresponding unshielded triples are marked as ambiguous. This is the case in the output $\Ehat$ of the conservative PC algorithm in Figure~\ref{fig:example1}. Crucially, the output $\Ehat$ is best possible in this example. Indeed, the given probability distribution could have also been generated by an SEM whose true DAG is consistent with $\widehat{C}$ (therefore entailing $X_1 \perp X_4$) and whose parameters are tuned to entail the faithfulness violation $X_1 \perp X_4 \mid X_2, X_3$. Therefore, the orientation of the ambiguous unshielded triples in $\Ehat$ is indeed not identifiable. 

However, the conservative PC algorithm still uses~\ref{design-choice-2} to restrict the number of conditional independence tests, which can lead to failures when adjacency faithfulness is violated (see Figure~\ref{fig:faithfulness} and Definition~\ref{defn:adjacency-faithfulness} in Appendix~\ref{apx:background}). In the example of Figure~\ref{fig:example2}, both the PC and the conservative PC algorithm delete the adjacencies between $X_1, X_3$ and $X_1, X_4$ early because of unfaithful independence. Note that conditioning on $X_1$ is required to find the only conditional independence between $X_3, X_4$, which is $X_3 \perp X_4 \mid X_1, X_2$. However, when $X_1$ is not adjacent to $X_3, X_4$ anymore, neither of the algorithms performs this test, and therefore both of them keep the false positive adjacency $\{X_3, X_4\}$. 

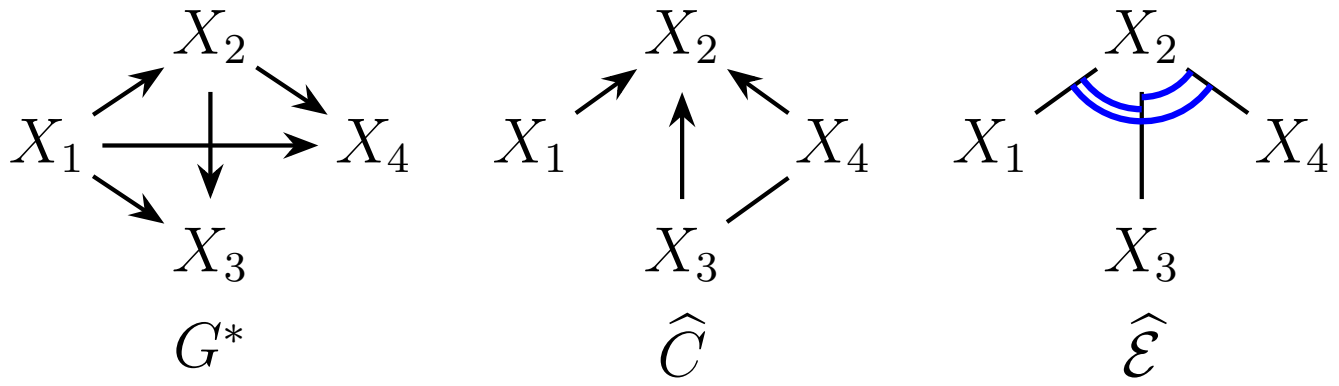
\begin{figure}
    \centering
\begin{tikzpicture}[scale=0.8,    >=Stealth,    node/.style={circle, inner sep=1pt, font=\large},    edge/.style={->, thick},  undirected/.style={-, thick},  lab/.style={font=\normalsize},
amb/.style={very thick, blue}]

\node[node] (L2) at (0,1.8) {$X_2$};
\node[node] (L1) at (-1.35,0.9) {$X_1$};
\node[node] (L4) at (1.35,0.9) {$X_4$};
\node[node] (L3) at (0,0) {$X_3$};

\node[node] (G1) at (0,-0.75) {$G^*$};

\draw[edge] (L1) -- (L2);
\draw[edge] (L1) -- (L3);
\draw[edge] (L1) -- (L4);
\draw[edge] (L2) -- (L3);
\draw[edge] (L2) -- (L4);

\node[node] (M2) at (3.9,1.8) {$X_2$};
\node[node] (M1) at (2.65,0.9) {$X_1$};
\node[node] (M4) at (5.15,0.9) {$X_4$};
\node[node] (M3) at (3.9,0) {$X_3$};

\node[node] (G2) at (3.9,-0.75) {$\widehat{C}$};

\draw[edge] (M1) -- (M2);
\draw[edge] (M3) -- (M2);
\draw[edge] (M4) -- (M2);
\draw[undirected] (M4) -- (M3);

\node[node] (R2) at (7.7,1.8) {$X_2$};
\node[node] (R1) at (6.45,0.9) {$X_1$};
\node[node] (R4) at (8.95,0.9) {$X_4$};
\node[node] (R3) at (7.7,0) {$X_3$};

\node[node] (G3) at (7.7,-0.75) {$\Ehat$};

\draw[undirected] (R1) -- (R2);
\draw[undirected] (R3) -- (R2);
\draw[undirected] (R4) -- (R2);

    \draw[amb] (7.7,1.8) ++(-35:0.5cm)
    arc[start angle=-35, end angle=-90, radius=0.5cm];
    \draw[amb] (7.7,1.8) ++(-35:0.7cm)
    arc[start angle=-35, end angle=-145, radius=0.7cm];
    \draw[amb] (7.7,1.8) ++(-90:0.6cm)
    arc[start angle=-90, end angle=-145, radius=0.6cm];

\end{tikzpicture}

    \caption{Consider a linear Gaussian SEM $(A, \mathbf{1})$ with corresponding DAG $G^*$ whose edge weights satisfy $A_{31} + A_{21}A_{32} = A_{41}+A_{21}A_{42} = 0$, implying the unfaithful independencies $X_1 \perp X_3$ and $X_1 \perp X_4$. With infinite samples, the PC and conservative PC algorithm output $\Chat$ with the false positive adjacency $\{X_3, X_4\}$, while the $m$-conservative PC algorithm correctly outputs $\Ehat$.}
    \label{fig:example2}
\end{figure}

To get an algorithm with error guarantees even in the absence of adjacency faithfulness, we need to replace paradigm~\ref{design-choice-2} with a strategy that allows us to test correct separating sets for each non-adjacent pair of vertices in the true DAG, ideally without performing exponentially many such tests. Our idea for the $m$-conservative PC algorithm (Algorithm~\ref{alg:main}) is to restrict the conditioning sets for each pair of vertices $\{i,j\}$ to the distance-$m$-neighborhood of $i$ and $j$ in the estimated undirected graph of the causal model (that we denote by $\Ne^m(i,\widehat{H})$ and  $\Ne^m(j,\widehat{H})$ respectively). To see that this works, consider the true \emph{parent separation statements} for a DAG on $[n]$, given by
\begin{align*}
    \cP(G) = \{(\{i,j\}, \Pa(j, G)) \mid i < j, i \notin \Pa(j,G)\}. 
\end{align*}
Each parent separation statement of $G$ must correspond to a conditional independence in a causal model with corresponding DAG $G$ by Lemma~\ref{lem:parentseparation} and the causal Markov condition (Proposition~\ref{prop:CMC}). Moreover, by the characterization of \citet{kocka_characterizing_2001} (Lemma~\ref{lem:char_inclusion}), any estimated DAG that entails the true parent separation statements does not have false positive error. Finally, we know by Proposition~\ref{prop:distance-m} that $\Pa(j,G) \subseteq \Ne^m(j, H)$ when $m$-connected-collider faithfulness holds. Hence, the $m$-conservative PC algorithm is guaranteed to test a correct parent separation statement for each pair of non-adjacent vertices in the true causal graph under this condition. In the example in Figure~\ref{fig:example2}, the undirected graph of the true causal model is the complete graph that is only missing $\{X_3, X_4\}$, hence the $m$-conservative PC algorithm for all $m \geq 1$ correctly tests $X_3 \perp X_4 \mid X_1, X_2$ and outputs the e-pattern $\Ehat$. Also in this example, $\Ehat$ is best possible, because the probability distribution of the causal model could have been generated by a DAG consistent with $\widehat{C}$ that has tuned weights to generate $X_3 \perp X_4 \mid X_1, X_2$ as an unfaithful conditional independency. Hence, the collider assignment of the unshielded triples in $\Ehat$ is indeed unidentifiable.

\begin{algorithm} 
    \caption{\textsc{$m$-Conservative PC Algorithm}} \label{alg:main}
     \begin{lstlisting}[style=pseudocode,mathescape,numbers=left, firstnumber=0, escapechar=|,columns=fullflexible,breaklines=true]
     Input: set of samples $\cS$ of $\bX = (X_1, \dots, X_n)$, significance level $\alpha$.
    $\widehat{H} \leftarrow$ |\textsc{UndirectedGraphLearning($\cS$)}|
    $\mathcal{L} \leftarrow \emptyset$
    for all $1 \leq i < j \leq n$: |\label{line-start-list}|
        for all $Z \subseteq \Ne^m(i, \widehat{H}) \setminus \{i,j\}$ and all $Z \subseteq \Ne^m(j, \widehat{H}) \setminus \{i,j\}$:
                $\mathcal{L} \leftarrow \mathcal{L} \cup \{(\{i,j\},Z)\}$ |\label{line-end-list}|
    $\Ihat \leftarrow$ |\textsc{TestConditionalIndependence($\cL, \cS, \alpha$)}| |\label{CI-line}|
    $\Ehat \leftarrow $ complete undirected graph on $[n]$
    for all $\{i,j\}$ such that there exists $Z$ with $(\{i,j\},Z) \in \Ihat$: |\label{start-skeleton}|
        delete $\{i,j\}$ from $\Ehat$ |\label{end-skeleton}|
    $\widehat{\mathcal{E}} \leftarrow$ |\textsc{OrientEPattern}($\Ehat, \Ihat$)|
    return $\widehat{\mathcal{E}}$
     \end{lstlisting}  
 \end{algorithm}

\subsection{Multiple-hypothesis testing}\label{sec:multiple-testing}
We test the selected conditional independence hypotheses, by assuming we have access to valid $p$-values. For Gaussian data, these can be approximated by Fisher’s $z$-test (see e.g. \citet{kalisch_estimating_2007}). In the context of constrained-based causal discovery, accepting any conditional independence usually removes an adjacency, and hence, conservative testing can result in very sparse graphs. Instead of trying to control family-wise error, we therefore target expected false discovery rate using a modified Benjamini–Yekutieli procedure (see Algorithm~\ref{alg:fdr} in Appendix~\ref{apx:proofs}) that accommodates dependent $p$-values~\citep{benjamini_control_2001}. Let us define the number of \emph{false positive parent statements} $\FPP(\Ihat, G^*)$ for a list $\Ihat$ of estimated conditional independencies and a true DAG $G^*$ and the number of \emph{induced adjacencies} $\Ad(\Ihat)$ as 
\begin{align*}
    \FPP(\Ihat, G^*) = \left| \cP(G^*) \setminus \Ihat \right|, \qquad  \Ad(\Ihat) = \left|\{\{i,j\} \mid \forall Z: (\{i,j\},Z) \notin\Ihat\}\right|.
\end{align*}
Indeed, according to lines~\ref{start-skeleton} and~\ref{end-skeleton} of Algorithm~\ref{alg:main}, $\Ad(\Ihat)$ is exactly the number of adjacencies of the final output. Our conditional independence testing procedure provides the following guarantee, which we prove in Appendix~\ref{apx:proofs}.

\begin{restatable}{prop}{fdrcontrol} \label{prop:fdr}
    Let $(A, \bv)$ be a linear Gaussian SEM for $n$ variables with corresponding DAG $G^* = G(A)$. If Algorithm~\ref{alg:fdr} has access to valid $p$-values and the input list $\cL$ contains $\cP(G^*)$, then 
    \begin{align*}
        \E\left[\frac{\FPP(\Ihat, G^*)}{\Ad(\Ihat)}\right] \leq \alpha.
    \end{align*}
\end{restatable}

\subsection{The orientation step} \label{sec:orientation-step}
After determining the induced adjacencies of $\Ihat$, the final step of the $m$-conservative PC algorithm is to orient these adjacencies and determine ambiguous unshielded triples. We state detailed pseudocode for the algorithm $\textsc{OrientEPattern}$ as Algorithm~\ref{alg:orient} in Appendix~\ref{apx:proofs} and give a high-level overview here: First, we apply the same rules as the conservative PC algorithm for marking unshielded triples as colliders, non-colliders or ambiguous. However, correctness of this orientation step requires not only adjacency faithfulness, but also oracle independence tests~\citep{ramsey_adjacency-faithfulness_2006}. Note that our algorithm claims an error guarantee only under the assumption of valid $p$-values for conditional independence tests. In particular, we do not assume any control on the type-II error rate of these tests, and therefore the $m$-conservative PC algorithm needs to robustly handle arbitrary false negative conditional independencies. We show this is possible with the following two additional rules:
\begin{enumerate}
    \item if some assignment of the ambiguous triples to either colliders or non-colliders conflicts with the assignment of an unambiguous triple, then we make it ambiguous;
    \item we make unshielded triples ambiguous that form a d-connecting path contradicting a tested conditional independency. 
\end{enumerate}
Figure~\ref{fig:example2} illustrates why the first step is necessary. After finding all correct conditional independencies, and choosing adjacencies accordingly, three edges and three unshielded triples remain. Based on their true separation sets, $(\{X_1, X_3\}, X_2)$ and $(\{X_1, X_4\}, X_2)$ are unambiguous colliders and $(\{X_3, X_4\}, X_2)$ is an unambiguous non-collider. Say, we also picked up the false conditional independence claim $X_3 \perp X_4$ due to a random test error. The conservative PC algorithm would now mark $(\{X_3, X_4\}, X_2)$ as ambiguous. However, the resulting e-pattern would only represent one valid DAG where $(\{X_3, X_4\}, X_2)$ is a collider and therefore incurs false positive error, since the edges $X_1 \to X_2, X_3 \to X_2, X_4 \to X_2$ are already forced by the unambiguous colliders. Hence, $(\{X_1, X_3\}, X_2)$ and $(\{X_1, X_4\}, X_2)$ must also be marked as ambiguous. 

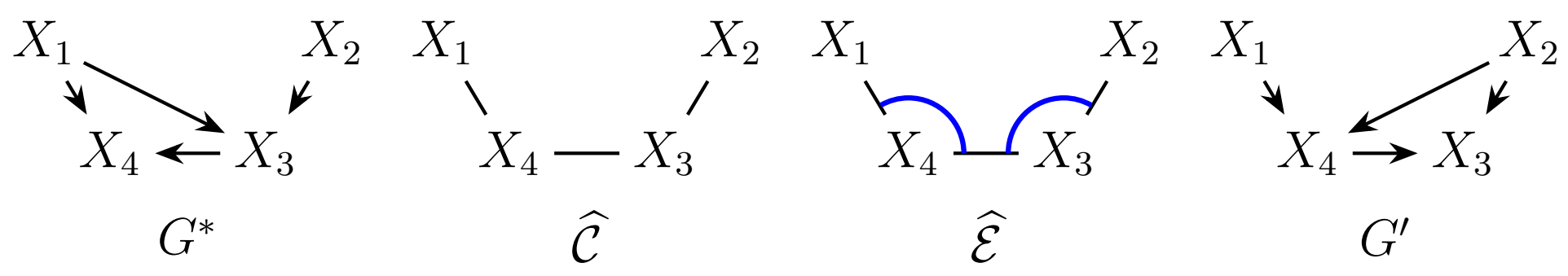
\begin{figure}
    \centering
\begin{tikzpicture}[
    >=Stealth,
    scale=0.9,
    node/.style={circle, inner sep=1pt, font=\large},
    edge/.style={->, thick},
    undirected/.style={-, thick},
    amb/.style={very thick, blue}
]

\node[node] (L1) at (0.0,1.0) {$X_1$};
\node[node] (L4) at (0.6,0.0) {$X_4$};
\node[node] (L3) at (2,0.0) {$X_3$};
\node[node] (L2) at (2.6,1.0) {$X_2$};

\node[node] at (1.3,-0.75) {$G^*$};

\draw[edge] (L1) -- (L4);
\draw[edge] (L1) -- (L3);
\draw[edge] (L2) -- (L3);
\draw[edge] (L3) -- (L4);

\node[node] (M1) at (3.6,1.0) {$X_1$};
\node[node] (M4) at (4.2,0.0) {$X_4$};
\node[node] (M3) at (5.6,0.0) {$X_3$};
\node[node] (M2) at (6.2,1.0) {$X_2$};

\node[node] at (4.9,-0.75) {$\Chat$};

\draw[undirected] (M1) -- (M4);
\draw[undirected] (M4) -- (M3);
\draw[undirected] (M3) -- (M2);

\node[node] (R1) at (7.2,1.0) {$X_1$};
\node[node] (R4) at (7.8,0.0) {$X_4$};
\node[node] (R3) at (9.2,0.0) {$X_3$};
\node[node] (R2) at (9.8,1.0) {$X_2$};

\node[node] at (8.5,-0.75) {$\Ehat$};

\draw[undirected] (R1) -- (R4);
\draw[undirected] (R4) -- (R3);
\draw[undirected] (R3) -- (R2);

\draw[amb] (7.8,0) ++(0:0.5cm)
    arc[start angle=0, end angle=120, radius=0.5cm];
\draw[amb] (9.2,0) ++(180:0.5cm)
    arc[start angle=180, end angle=60, radius=0.5cm];

\node[node] (RR1) at (10.8,1.0) {$X_1$};
\node[node] (RR4) at (11.4,0.0) {$X_4$};
\node[node] (RR3) at (12.8,0.0) {$X_3$};
\node[node] (RR2) at (13.4,1.0) {$X_2$};

\draw[edge] (RR1) -- (RR4);
\draw[edge] (RR2) -- (RR3);
\draw[edge] (RR2) -- (RR4);
\draw[edge] (RR4) -- (RR3);

\node[node] at (12.1,-0.75) {$G'$};

\end{tikzpicture}
    \caption{Suppose the linear Gaussian SEM $(A, \mathbf{1})$ with corresponding DAG $G^*$ satisfies $A_{31} = A_{41}A_{43}(A_{32}^2 + 1)$, implying the unfaithful independence $X_1 \perp X_3 \mid X_4$. With infinite samples, the PC and conservative PC algorithm output $\Chat$ with the false positive d-connection $X_1 \not \perp X_2$, while the $m$-conservative PC algorithm correctly outputs $\Ehat$. The distribution of the true model could have also been generated by $G'$ with unfaithful independency $X_2 \perp X_4 \mid X_3$.}
    \label{fig:example3}
\end{figure}

To motivate the second step, consider Figure~\ref{fig:example3}. All of the unshielded triples only have separating sets that render them unambiguous non-colliders. Even though the PC and the conservative PC algorithm test $X_1 \perp X_2$, no step ensures that the output CPDAG still entails this test as a d-separation. However, the \textsc{OrientEPattern} algorithm tests for this contradiction and makes both triples in the graph ambiguous. Again, the resulting e-pattern $\Ehat$ is the best possible output because the true distribution could have been generated by $G'$, which makes the triple orientations unidentifiable. 

Together, both steps above ensure that even when the estimated list $\Ihat$ of conditional independencies in the $m$-conservative PC algorithm only contains a subset of the true parent separation statements $\Phat \subseteq \cP(G^*)$, the final e-pattern represents a DAG that entails all separation statements in $\Phat$ after adjusting for false positive adjacencies. In particular, the output of \textsc{OrientEPattern} always represents at least one valid DAG for any arbitrary list of conditional independencies $\Ihat$, which means that the output of the $m$-conservative PC algorithm always has a causal interpretation whereas the PC or conservative PC algorithm may output partially directed graphs with no valid DAG extensions in practice. 

\begin{restatable}{prop}{orientation} \label{prop:CIstatements}
    Let $S$ be an undirected graph and $\cL$ be a list of conditional independence relations over a vertex set $V$. Let $\cA \subseteq \cL$ be a subset of conditional independence relations that are also entailed as d-separations by a DAG $G^*$ on $V$. Then, the output $\widehat{\cE}$ of \textsc{OrientEPattern}($S, \cL$) represents a DAG $\widehat{G}$ with the following property: let $U$ be any vertex cover for the set of edges in $\widehat{G}$ between non-adjacent vertices in $G^*$. Then, deleting all edges incident to $U$ results in a DAG $\widehat{G}'$ that entails all the conditional independence statements in $\cA$ as d-separations. 
\end{restatable}

Propositions~\ref{prop:fdr} and~\ref{prop:CIstatements} together imply Theorem~\ref{thm:main}. We refer to Appendix~\ref{apx:proofs} for the full proof. 
 
\paragraph{Complexity analysis.}
The main bottlenecks of the $m$-conservative PC algorithm are the number of conditional independence tests and the runtime of \textsc{OrientEPattern}. The number of conditional independence tests is bounded by $\binom{n}{2}2^{d^m+1}$, where $d$ is the degree of the estimated undirected graph $\widehat{H}$. When this is the same as the degree of the true undirected graph $H^*$ and the true DAG $G^*$ has maximum degree $\Delta$, then $d$ is bounded by $\Delta^2$ (see Lemma~\ref{lem:degree} in Appendix~\ref{apx:proofs}). This shows the trade-off between larger $m$, which allows the algorithm to handle significantly more faithfulness violations, and a double-exponential increase in runtime. As a special case, note that when $m$ is large enough, then the $m$-conservative PC algorithm eventually performs all $\binom{n}{2}2^{n-2}$ possible conditional independence tests, and makes an error guarantee without any faithfulness assumption as $m$-connected-collider faithfulness becomes vacuous. The dominating factor in the runtime of \textsc{OrientEPattern} is the number of final ambiguities $a$. Since \textsc{OrientEPattern} has to check all possible assignments to the ambiguous triples, its runtime scales exponentially in $a$. This shows that the $m$-conservative PC algorithm becomes intractable if there are many empirical faithfulness violations or the true DAG has large degree, which is a somewhat expected price to pay for the strong error guarantees given that DAG learning is \textrm{NP}-complete~\citep{chickering_learning_1996}. In Table~\ref{tab:mcpc-scalability}, Appendix~\ref{apx:experiments}, we provide an empirical runtime analysis for the $m$-conservative PC algorithm.

\section{Experiments} \label{sec:experiments}

\begin{figure}
    \centering

    \begin{subfigure}{0.49\textwidth}
        \centering
        \includegraphics[width=\linewidth]{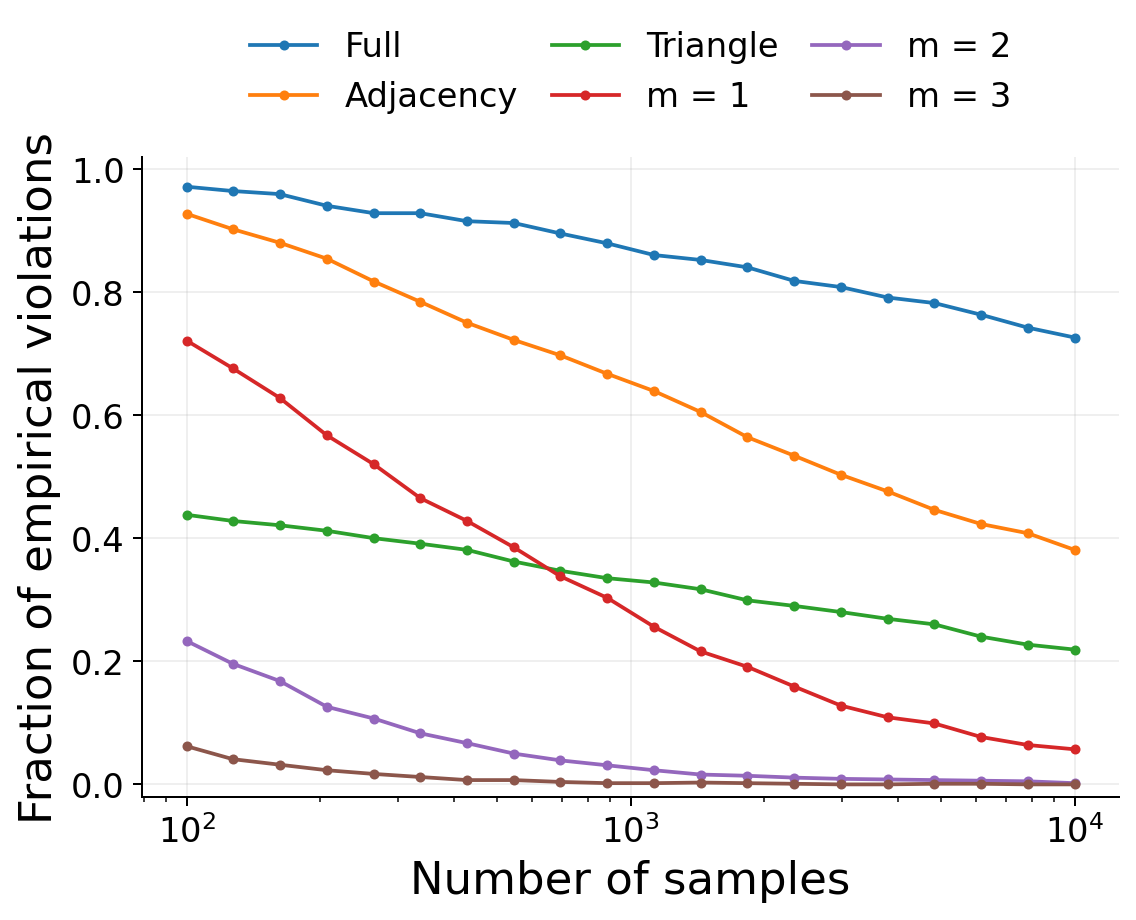}
        \caption{Empirical faithfulness violations.}
        \label{fig:faithfulness}
    \end{subfigure}
    \hfill
    \begin{subfigure}{0.49\textwidth}
        \centering
        \includegraphics[width=\linewidth]{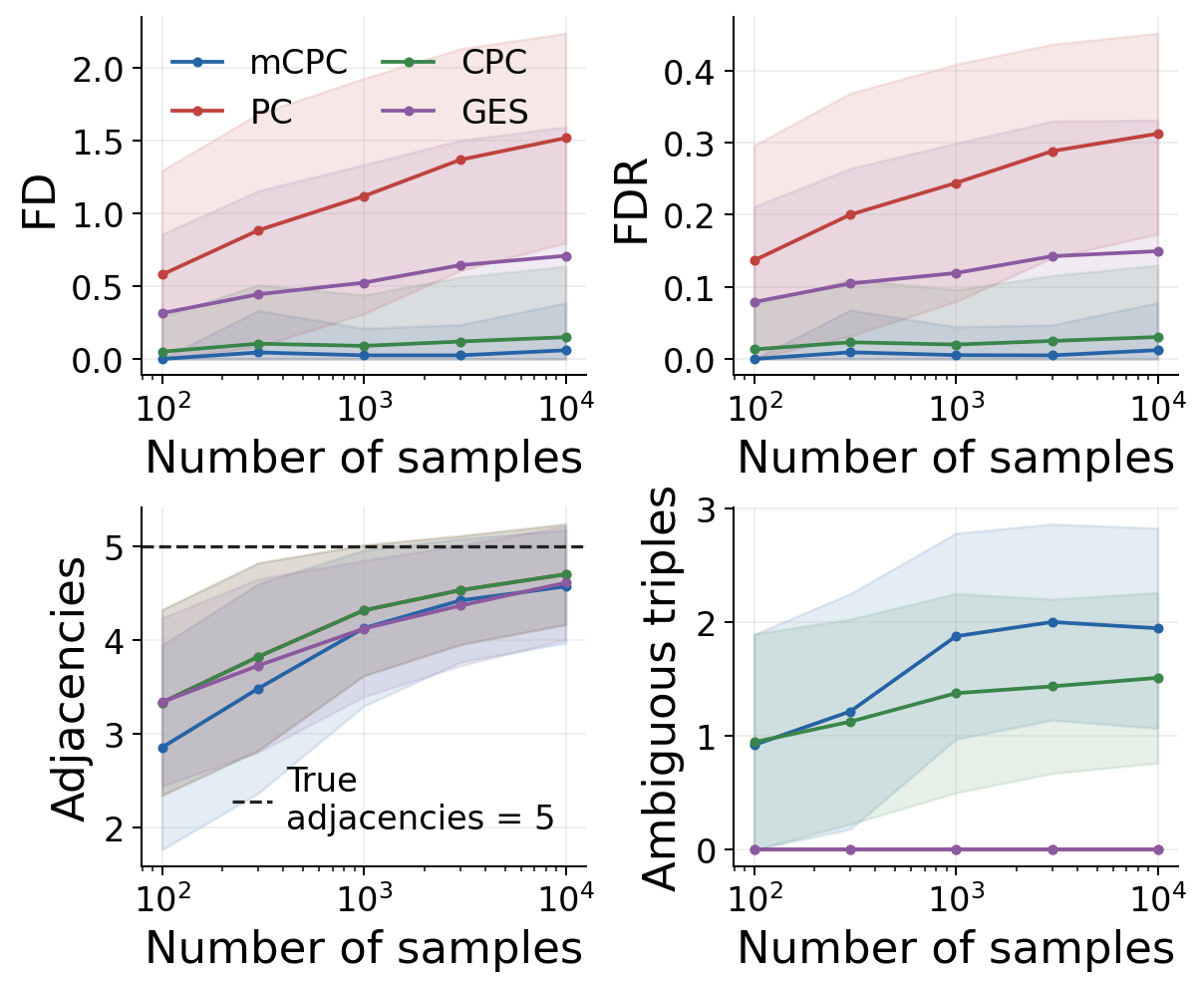}
        \caption{Results for the model in Figure~\ref{fig:example1}.}
        \label{fig:first}
    \end{subfigure}
    \hfill
    \begin{subfigure}{0.49\textwidth}
        \centering
        \includegraphics[width=\linewidth]{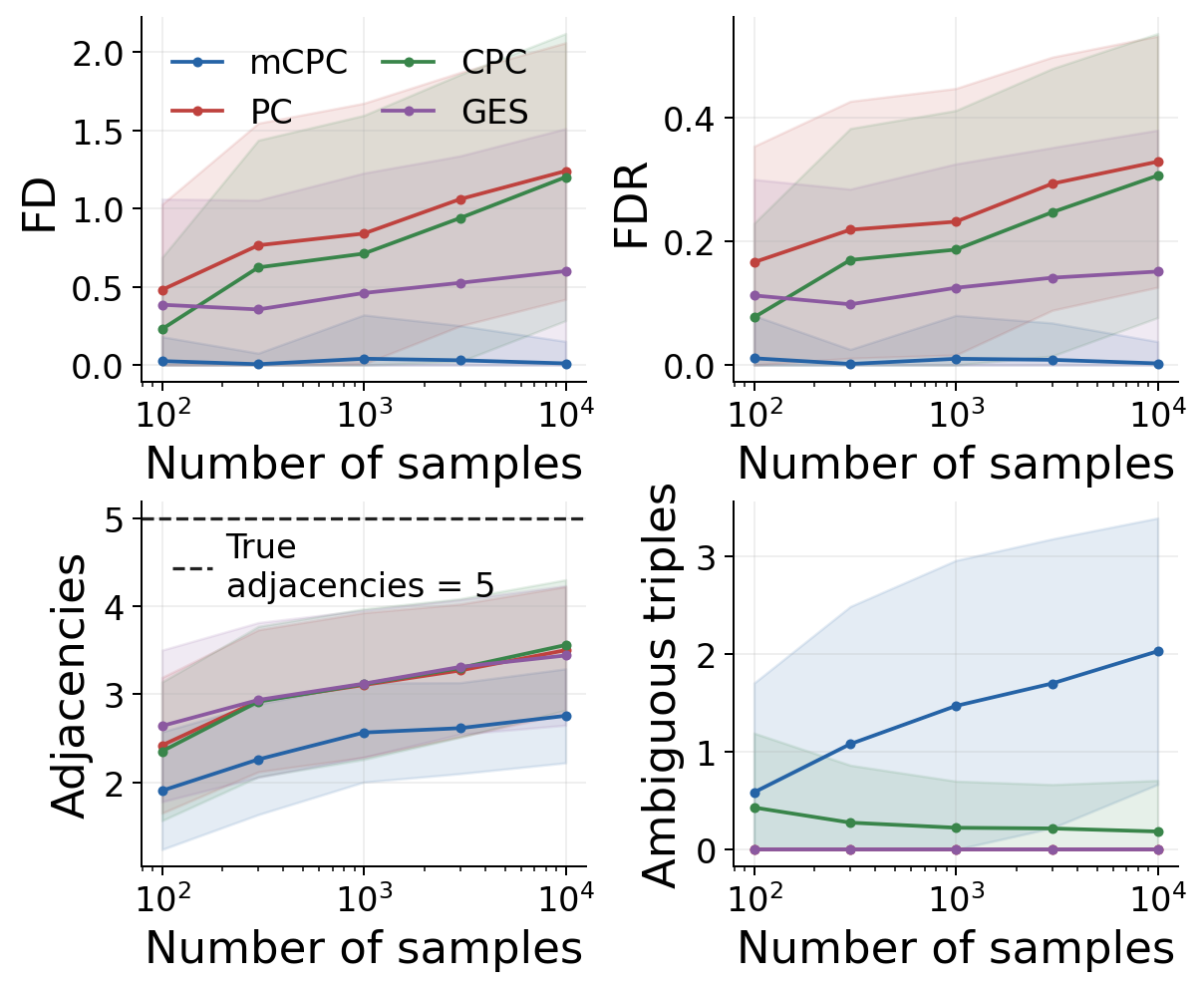}
        \caption{Results for the model in Figure~\ref{fig:example2}.}
        \label{fig:second}
    \end{subfigure}
    \hfill
    \begin{subfigure}{0.49\textwidth}
        \centering
        \includegraphics[width=\linewidth]{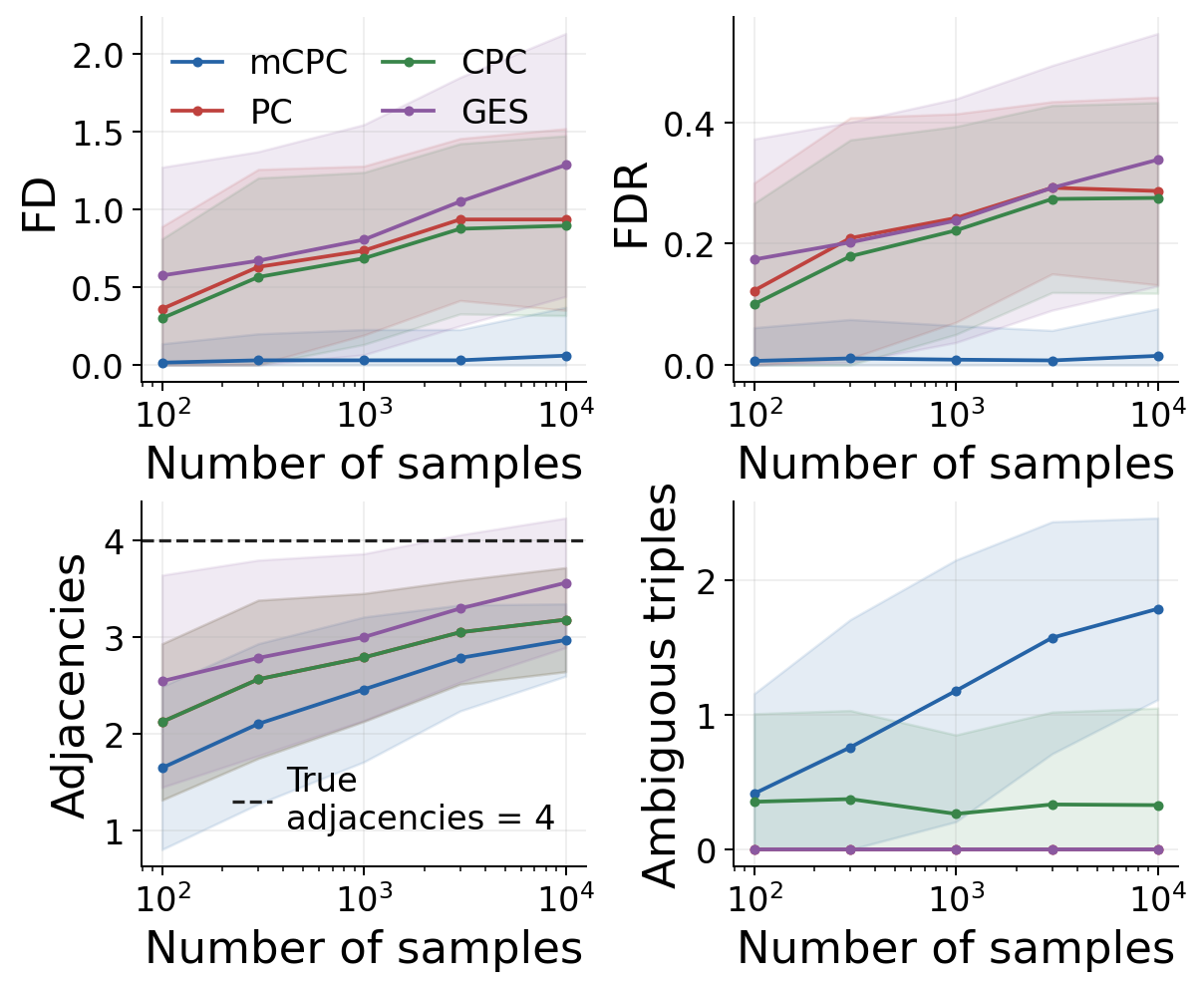}
        \caption{Results for the model in Figure~\ref{fig:example3}.}
        \label{fig:third}
    \end{subfigure}

    \caption{Experimental results for faithfulness violations. Shaded areas correspond to standard deviations over multiple randomly sampled models.}
    \label{fig:experiments}
\end{figure}

First, we test the strength of the different faithfulness conditions empirically. While the set of parameters violating any faithfulness condition has measure zero, near-violations can be common in practice, which makes the difference between $m$-connected-collider faithfulness and other assumptions significant. We randomly sampled $1000$ linear Gaussian SEMs on $10$ variables (see Appendix~\ref{apx:experiments} for details) and empirically estimated the conditional independencies of each model, using Fisher's $z$-test with significance level $\alpha = 0.1$. We compared the results to the d-separation relations entailed by the true DAG to determine faithfulness violations. Figure~\ref{fig:faithfulness} shows that empirical faithfulness violations can be very common (which is consistent with previous research~\citep{uhler_geometry_2013}), but already for $m=2$, empirical violations of $m$-connected-collider faithfulness are rare in our setting.

\begin{figure}
    \centering
    \begin{subfigure}{0.49\textwidth}
        \centering 
        \includegraphics[width=\linewidth]{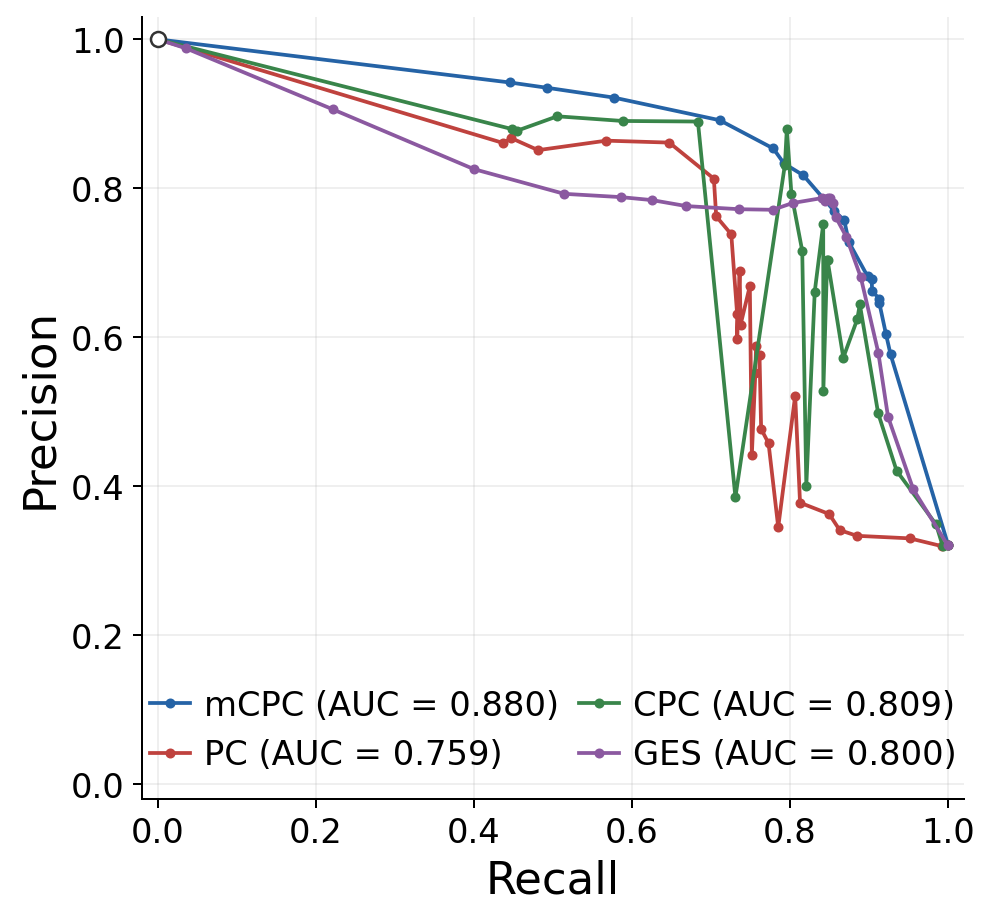}
        \caption{Results for randomly sampled ground truth models.}
        \label{fig:pr-random}
    \end{subfigure}
    \hfill 
    \begin{subfigure}{0.49\textwidth}
        \centering 
        \includegraphics[width=\linewidth]{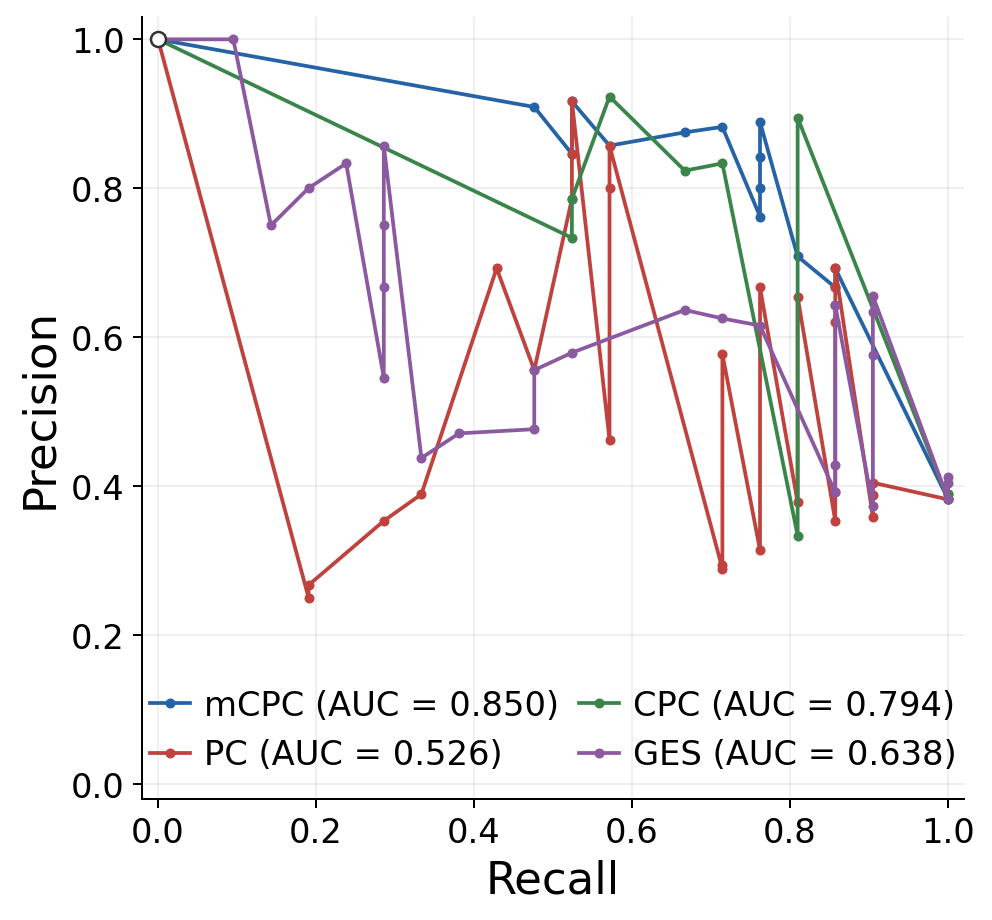}
        \caption{Results for the Causal Chamber dataset.}
        \label{fig:pr-chambers}
    \end{subfigure}
    \caption{Precision-Recall curves.}
\end{figure}

\begin{figure}
    \centering
    \includegraphics[width=\linewidth]{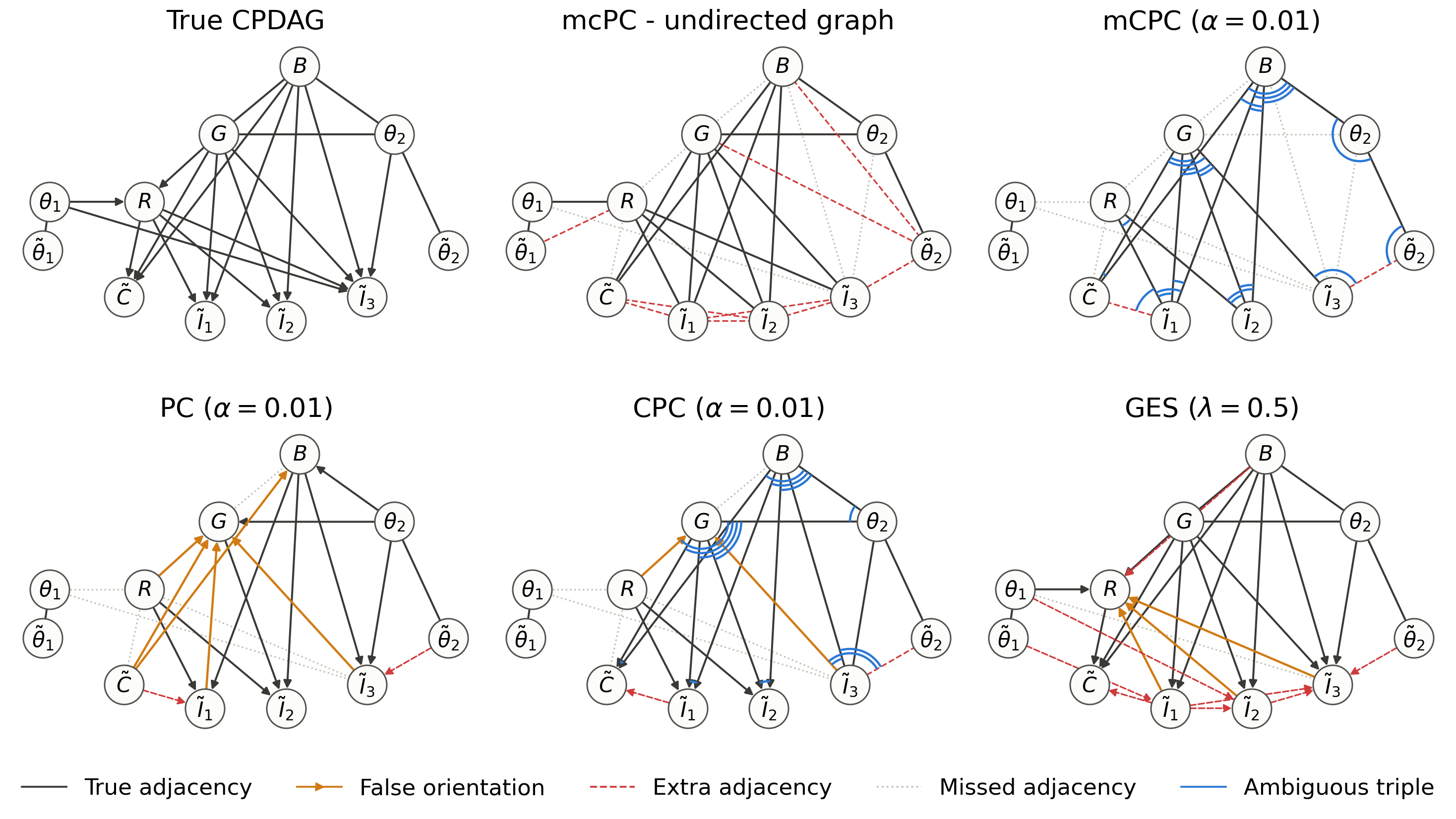}
    \caption{Ground truth and output graphs of the $m$-conservative PC algorithm, its intermediate undirected graph learning step, and the baseline algorithms for the Causal Chamber dataset.}
    \label{fig:graphs}
\end{figure}

Furthermore, we implemented the $m$-conservative PC algorithm and tested it on the following datasets: First, we sampled synthetic data according to the unfaithful linear  Gaussian SEMs described in Figures~\ref{fig:example1},~\ref{fig:example2} and~\ref{fig:example3} for different numbers of samples. As expected, the $m$-conservative PC algorithm keeps the false discovery essentially at zero whereas the baseline algorithms PC, CPC and also GES~\citep{Chickering03} incur errors, see Figure~\ref{fig:experiments}. Next, we report Precision-Recall curves averaged over algorithm runs on 100 random linear Gaussian SEMs with 10 variables, expected degree 3 and 1000 samples each, see Figure~\ref{fig:pr-random}. Finally, we run the algorithms on 10000 samples from the Causal Chambers light tunnel with 11 variables, which has a known ground truth DAG~\citep{gamella_sanity_2025}, see Figures~\ref{fig:pr-chambers} and~\ref{fig:graphs}. Throughout all experiments, we use $m=1$ for our algorithm, which we recommend as the practical default setting. Our algorithm is competitive in terms of AUC with the baseline algorithms. However, it often marks many unshielded triples as ambiguous (Figure~\ref{fig:graphs}). This avoids orientation errors, but also makes the output less informative. If the goal is to infer a unique DAG, the output of our algorithm should be interpreted as a starting point, providing the edge information that can be reliably inferred from observational data only. Then, the e-pattern could be used to guide interventions to further orient edges, for instance intervening on nodes $B$ and $G$ in Figure~\ref{fig:graphs} would resolve most of the ambiguities. Implementation details for all experiments can be found in Appendix~\ref{apx:experiments} and our code is publicly available under \url{https://github.com/ejahn17/causal-FPE-control}.

\section{Discussion}\label{sec:discussion}
In this paper, we introduced the $m$-conservative PC algorithm and proved that it controls the expected false discovery rate. This guarantee necessarily relies on limiting assumptions. While we argue that $m$-connected-collider faithfulness is already much weaker than most faithfulness assumptions, we still assume a linear Gaussian setting, which is likely to be violated in practice. We remark that the only part of our approach that requires this assumption is the connection between $m$-connected-collider faithfulness and the undirected graph of the causal model (Proposition~\ref{prop:distance-m}). This is needed to show that the algorithm selects sufficiently many conditional independence tests. The testing procedure itself only relies on valid $p$-values, which can also be obtained for other classes of distributions, and the final orientation step is purely combinatorial. Future work could extend our strategy for selecting conditional independence tests by a method that works beyond the linear Gaussian setting. 

Another limitation of our approach is that the $m$-conservative PC algorithm outputs an e-pattern, which comes with two drawbacks: first, it can mark many edge directions as ambiguous and therefore represent a large number of DAGs, which makes the output hard to interpret. Second, the runtime of the algorithm scales exponentially in the number of ambiguities, which makes it infeasible for high-dimensional data. This might be related to fundamental limitations of causal discovery in the absence of faithfulness: We showed multiple examples where the ambiguous triples in the output of the $m$-conservative PC algorithm are genuinely unidentifiable. We hope that these findings can spark further research that formalizes lower bounds on what can still be learned from observational data under conditions such as $m$-connected-collider faithfulness. 

Rather than aiming to infer a unique DAG, we view our algorithm as useful in settings where one only wants to extract the partial causal information that is most strongly supported by the data, especially when false positive causal claims could lead to potentially harmful consequences. Moreover, our algorithm can be seen as a useful first step in experiment design: the orientation of the edges of each ambiguous triple in an e-pattern can be resolved by a single intervention on the center vertex of that triple. Hence, the output of the $m$-conservative PC algorithm can be used to guide a minimal set of interventions that, combined with the unambiguous findings from the observational data, recovers a full DAG with controlled false positive error. 

\begin{ack}
    This research was supported by NSF grant CCF-2321079.
\end{ack}

\bibliographystyle{plainnat}
\bibliography{refs}

@article{
TBC24,
author = {Armeen Taeb  and Peter Bühlmann  and Venkat Chandrasekaran },
title = {Model selection over partially ordered sets},
journal = {Proceedings of the National Academy of Sciences},
volume = {121},
number = {8},
pages = {e2314228121},
year = {2024},
URL = {https://www.pnas.org/doi/abs/10.1073/pnas.2314228121},
}

@article{Chickering03,
	title = {Optimal {Structure} {Identification} {With} {Greedy} {Search}},
	volume = {3},
	issn = {1533-7928},
	url = {https://jmlr.org/papers/v3/chickering02b.html},
	number = {Nov},
	urldate = {2026-02-22},
	journal = {Journal of Machine Learning Research},
	author = {Chickering, David Maxwell},
	year = {2002},
	pages = {507--554},
}

@article{MB06,
author = {Meinshausen, Nicolai and B{\"u}hlmann, Peter},
title = {{High-dimensional graphs and variable selection with the Lasso}},
volume = {34},
journal = {The Annals of Statistics},
number = {3},
publisher = {Institute of Mathematical Statistics},
pages = {1436 -- 1462},
year = {2006},
URL = {https://doi.org/10.1214/009053606000000281}
}

@inproceedings{ramsey_adjacency-faithfulness_2006,
	author = {Ramsey, Joseph and Spirtes, Peter and Zhang, Jiji},
	address = {Arlington, Virginia, USA},
	series = {{UAI}'06},
	title = {Adjacency-faithfulness and conservative causal inference},
	isbn = {978-0-9749039-2-7},
	urldate = {2026-03-18},
	booktitle = {Proceedings of the {Twenty}-{Second} {Conference} on {Uncertainty} in {Artificial} {Intelligence}},
	publisher = {AUAI Press},
	year = {2006},
	pages = {401--408},
}

@inproceedings{kocka_characterizing_2001,
	address = {San Francisco, CA, USA},
	series = {{UAI}'01},
	title = {On characterizing inclusion of {Bayesian} networks},
	isbn = {978-1-55860-800-9},
	url = {https://dl.acm.org/doi/10.5555/2074022.2074055},
	booktitle = {Proceedings of the {Seventeenth} conference on {Uncertainty} in artificial intelligence},
	publisher = {Morgan Kaufmann Publishers Inc.},
	author = {Kočka, Tomáš and Bouckaert, Remco R. and Studeny, Milan},
	month = aug,
	year = {2001},
	pages = {261--268},
}

@inproceedings{verma_equivalence_1990,
	address = {USA},
	series = {{UAI} '90},
	title = {Equivalence and synthesis of causal models},
	isbn = {978-0-444-89264-5},
	urldate = {2026-04-27},
	booktitle = {Proceedings of the {Sixth} {Annual} {Conference} on {Uncertainty} in {Artificial} {Intelligence}},
	publisher = {Elsevier Science Inc.},
	author = {Verma, Thomas and Pearl, Judea},
	month = jul,
	year = {1990},
	pages = {255--270},
}

@article{loh_high-dimensional_2014,
	title = {High-dimensional learning of linear causal networks via inverse covariance estimation},
	volume = {15},
	issn = {1532-4435},
	url = {https://dl.acm.org/doi/10.5555/2627435.2697063},
	number = {1},
	journal = {J. Mach. Learn. Res.},
	author = {Loh, Po-Ling and Bühlmann, Peter},
	month = jan,
	year = {2014},
	pages = {3065--3105},
}

@article{benjamini_control_2001,
	title = {The control of the false discovery rate in multiple testing under dependency},
	volume = {29},
	issn = {0090-5364, 2168-8966},
	url = {https://projecteuclid.org/journals/annals-of-statistics/volume-29/issue-4/The-control-of-the-false-discovery-rate-in-multiple-testing/10.1214/aos/1013699998.full},
	number = {4},
	urldate = {2026-04-30},
	journal = {The Annals of Statistics},
	publisher = {Institute of Mathematical Statistics},
	author = {Benjamini, Yoav and Yekutieli, Daniel},
	month = aug,
	year = {2001},
	pages = {1165--1188},
}

@misc{wang_elementary_2025,
	title = {Elementary proofs of several results on false discovery rate},
	url = {http://arxiv.org/abs/2201.09350},
	publisher = {arXiv},
	author = {Wang, Ruodu},
	month = jul,
	year = {2025},
	note = {arXiv:2201.09350 [math]},
}

@inproceedings{wienobst_extendability_2021,
	title = {Extendability of causal graphical models: {Algorithms} and computational complexity},
	issn = {2640-3498},
	shorttitle = {Extendability of causal graphical models},
	url = {https://proceedings.mlr.press/v161/wienobst21a.html},
	booktitle = {Proceedings of the {Thirty}-{Seventh} {Conference} on {Uncertainty} in {Artificial} {Intelligence}},
	publisher = {PMLR},
	author = {Wienöbst, Marcel and Bannach, Max and Liśkiewicz, Maciej},
	month = dec,
	year = {2021},
	pages = {1248--1257},
}

@inproceedings{meek_causal_1995,
	address = {San Francisco, CA, USA},
	series = {{UAI}'95},
	title = {Causal inference and causal explanation with background knowledge},
	isbn = {978-1-55860-385-1},
	url = {https://dl.acm.org/doi/10.5555/2074158.2074204},
	booktitle = {Proceedings of the {Eleventh} conference on {Uncertainty} in artificial intelligence},
	publisher = {Morgan Kaufmann Publishers Inc.},
	author = {Meek, Christopher},
	month = aug,
	year = {1995},
	pages = {403--410},
}

@article{zhang_detection_2008,
	title = {Detection of {Unfaithfulness} and {Robust} {Causal} {Inference}},
	volume = {18},
	issn = {1572-8641},
	url = {https://doi.org/10.1007/s11023-008-9096-4},
	number = {2},
	urldate = {2026-05-05},
	journal = {Minds and Machines},
	author = {Zhang, Jiji and Spirtes, Peter},
	month = jun,
	year = {2008},
	pages = {239--271},
}

@inproceedings{ng_reliable_2021,
	title = {Reliable {Causal} {Discovery} with {Improved} {Exact} {Search} and {Weaker} {Assumptions}},
	volume = {34},
	url = {https://proceedings.neurips.cc/paper/2021/hash/a9b4ec2eb4ab7b1b9c3392bb5388119d-Abstract.html},
	booktitle = {Advances in {Neural} {Information} {Processing} {Systems}},
	publisher = {Curran Associates, Inc.},
	author = {Ng, Ignavier and Zheng, Yujia and Zhang, Jiji and Zhang, Kun},
	year = {2021},
	pages = {20308--20320},
}

@article{uhler_geometry_2013,
	title = {Geometry of the {Faithfulness} {Assumption} in {Causal} {Inference}},
	volume = {41},
	issn = {0090-5364},
	url = {https://www.jstor.org/stable/23566569},
	number = {2},
	journal = {The Annals of Statistics},
	publisher = {Institute of Mathematical Statistics},
	author = {Uhler, Caroline and Raskutti, Garvesh and Bühlmann, Peter and Yu, Bin},
	year = {2013},
	pages = {436--463},
}

@article{kalisch_estimating_2007,
	title = {Estimating {High}-{Dimensional} {Directed} {Acyclic} {Graphs} with the {PC}-{Algorithm}},
	volume = {8},
	issn = {1533-7928},
	url = {http://jmlr.org/papers/v8/kalisch07a.html},
	number = {22},
	journal = {Journal of Machine Learning Research},
	author = {Kalisch, Markus and Bühlmann, Peter},
	year = {2007},
	pages = {613--636},
}

@book{spirtes_causation_2001,
	title = {Causation, {Prediction}, and {Search}},
	isbn = {978-0-262-28415-8},
	url = {https://direct.mit.edu/books/monograph/2057/Causation-Prediction-and-Search},
	language = {en},
	urldate = {2026-05-06},
	publisher = {The MIT Press},
	author = {Spirtes, Peter and Glymour, Clark and Scheines, Richard},
	month = jan,
	year = {2001},
}

@article{kelly_review_2022,
	title = {A review of causal discovery methods for molecular network analysis},
	volume = {10},
	issn = {2324-9269},
	url = {https://pmc.ncbi.nlm.nih.gov/articles/PMC9544222/},
	number = {10},
	journal = {Molecular Genetics \& Genomic Medicine},
	author = {Kelly, Jack and Berzuini, Carlo and Keavney, Bernard and Tomaszewski, Maciej and Guo, Hui},
	month = sep,
	year = {2022},
	pages = {e2055},
}

@article{jin_causal_2025,
	title = {Causal {Discovery} in {Astrophysics}: {Unraveling} {Supermassive} {Black} {Hole} and {Galaxy} {Coevolution}},
	volume = {979},
	issn = {0004-637X},
	shorttitle = {Causal {Discovery} in {Astrophysics}},
	url = {https://doi.org/10.3847/1538-4357/ad9ded},
	number = {2},
	journal = {The Astrophysical Journal},
	publisher = {The American Astronomical Society},
	author = {Jin, Zehao and Pasquato, Mario and Davis, Benjamin L. and Deleu, Tristan and Luo, Yu and Cho, Changhyun and Lemos, Pablo and Perreault-Levasseur, Laurence and Bengio, Yoshua and Kang, Xi and Macciò, Andrea Valerio and Hezaveh, Yashar},
	month = jan,
	year = {2025},
	pages = {212},
}

@article{addo_exploring_2021,
	title = {Exploring nonlinearity on the {CO}$_2$ emissions, economic production and energy use nexus: {A} causal discovery approach},
	volume = {7},
	issn = {2352-4847},
	url = {https://www.sciencedirect.com/science/article/pii/S2352484721008313},
	journal = {Energy Reports},
	author = {Addo, Peter Martey and Manibialoa, Christelle and McIsaac, Florent},
	month = nov,
	year = {2021},
	pages = {6196--6204},
}

@article{li_controlling_2009,
	title = {Controlling the {False} {Discovery} {Rate} of the {Association}/{Causality} {Structure} {Learned} with the {PC} {Algorithm}},
	volume = {10},
	issn = {1533-7928},
	url = {http://jmlr.org/papers/v10/li09a.html},
	number = {17},
	journal = {Journal of Machine Learning Research},
	author = {Li, Junning and Wang, Z. Jane},
	year = {2009},
	pages = {475--514},
}

@article{strobl_estimating_2019,
	title = {Estimating and {Controlling} the {False} {Discovery} {Rate} of the {PC} {Algorithm} {Using} {Edge}-specific {P}-{Values}},
	volume = {10},
	issn = {2157-6904},
	url = {https://dl.acm.org/doi/10.1145/3351342},
	number = {5},
	journal = {ACM Trans. Intell. Syst. Technol.},
	author = {Strobl, Eric V. and Spirtes, Peter L. and Visweswaran, Shyam},
	month = oct,
	year = {2019},
	pages = {46:1--46:37},
}

@article{benjamini_controlling_1995,
	title = {Controlling the {False} {Discovery} {Rate}: {A} {Practical} and {Powerful} {Approach} to {Multiple} {Testing}},
	volume = {57},
	issn = {0035-9246},
	shorttitle = {Controlling the {False} {Discovery} {Rate}},
	url = {https://www.jstor.org/stable/2346101},
	number = {1},
	journal = {Journal of the Royal Statistical Society. Series B (Methodological)},
	publisher = {[Royal Statistical Society, Oxford University Press]},
	author = {Benjamini, Yoav and Hochberg, Yosef},
	year = {1995},
	pages = {289--300},
}

@article{lemeire_conservative_2012,
	series = {Fifth {European} {Workshop} on {Probabilistic} {Graphical} {Models} ({PGM}-2010)},
	title = {Conservative independence-based causal structure learning in absence of adjacency faithfulness},
	volume = {53},
	issn = {0888-613X},
	url = {https://www.sciencedirect.com/science/article/pii/S0888613X12000801},
	number = {9},
	journal = {International Journal of Approximate Reasoning},
	author = {Lemeire, Jan and Meganck, Stijn and Cartella, Francesco and Liu, Tingting},
	month = dec,
	year = {2012},
	pages = {1305--1325},
}

@article{spirtes_uniformly_2014,
	title = {A {Uniformly} {Consistent} {Estimator} of {Causal} {Effects} under the $k$-{Triangle}-{Faithfulness} {Assumption}},
	volume = {29},
	issn = {0883-4237, 2168-8745},
	url = {https://projecteuclid.org/journals/statistical-science/volume-29/issue-4/A-Uniformly-Consistent-Estimator-of-Causal-Effects-under-the-k/10.1214/13-STS429.full},
	language = {en},
	number = {4},
	journal = {Statistical Science},
	publisher = {Institute of Mathematical Statistics},
	author = {Spirtes, Peter and Zhang, Jiji},
	month = nov,
	year = {2014},
	pages = {662--678},
}

@article{zhalama_sat-based_2017,
	title = {{SAT}-based causal discovery under weaker assumptions},
	url = {https://www.scopus.com/pages/publications/85031124700},
	journal = {Proceedings of the 33rd Conference on Uncertainty in Artificial Intelligence (UAI)},
	publisher = {Association for Uncertainty in Artificial Intelligence (AUAI)},
	author = {Zhalama and Zhang, Jiji and Eberhardt, Frederick and Mayer, Wolfgang},
	month = jan,
	year = {2017},
}

@book{pearl_causality_2009,
	address = {Cambridge},
	edition = {2},
	title = {Causality},
	isbn = {978-0-521-89560-6},
	url = {https://www.cambridge.org/core/books/causality/B0046844FAE10CBF274D4ACBDAEB5F5B},
	publisher = {Cambridge University Press},
	author = {Pearl, Judea},
	year = {2009},
}

@incollection{chickering_learning_1996,
	address = {New York, NY},
	title = {Learning {Bayesian} {Networks} is {NP}-{Complete}},
	isbn = {978-1-4612-2404-4},
	url = {https://doi.org/10.1007/978-1-4612-2404-4_12},
	booktitle = {Learning from {Data}: {Artificial} {Intelligence} and {Statistics} {V}},
	publisher = {Springer},
	author = {Chickering, David Maxwell},
	editor = {Fisher, Doug and Lenz, Hans-J.},
	year = {1996},
	pages = {121--130},
}

@article{gamella_causal_2025,
	title = {Causal chambers as a real-world physical testbed for {AI} methodology},
	volume = {7},
	copyright = {2025 The Author(s)},
	issn = {2522-5839},
	url = {https://www.nature.com/articles/s42256-024-00964-x},
	number = {1},
	journal = {Nature Machine Intelligence},
	publisher = {Nature Publishing Group},
	author = {Gamella, Juan L. and Peters, Jonas and Bühlmann, Peter},
	month = jan,
	year = {2025},
	pages = {107--118},
}

@inproceedings{gamella_sanity_2025,
	title = {Sanity {Checking} {Causal} {Representation} {Learning} on a {Simple} {Real}-{World} {System}},
	issn = {2640-3498},
	url = {https://proceedings.mlr.press/v267/gamella25a.html},
	booktitle = {Proceedings of the 42nd {International} {Conference} on {Machine} {Learning}},
	publisher = {PMLR},
	author = {Gamella, Juan L. and Bing, Simon and Runge, Jakob},
	month = oct,
	year = {2025},
	pages = {18143--18169},
}

\newpage
\appendix

\section{Additional background} \label{apx:background}
\textbf{Definitions and notation for graphical models.} A \emph{partially directed graph} $G$ consists of a set of vertices $V(G)$, a set of undirected edges $E(G)$, which are subsets $\{v,w\} \subseteq V$ of size $2$, and a set of directed edges (arcs) $A(G)$, which are ordered tuples $(v,w) \in V^2$ of size $2$ (we also denote a directed edge by $v \to w$). We only consider graphs without double edges and loops, that is at most one of the possible edges $\{v,w\}, (v,w), (w,v)$ is included in $E \cup A$ for all $v,w \in V$ and for all $v\in V$, we have $(v,v) \notin A$. Two vertices $v \neq w \in V$ are \emph{adjacent} if there exists a directed or undirected edge between them. We denote the number of adjacencies of $G$ by $\Ad(G) = |E(G)| + |A(G)|$. A \emph{path} in $G$ of length $n$ is an ordered list of vertices $(v_1, \dots, v_{n+1})$ such that $v_i$ and $v_{i+1}$ are adjacent for all $i = 1, \dots, n$. A path $(v_1, \dots, v_{n+1})$ is \emph{directed} if $(v_i, v_{i+1}) \in A(G)$ for all $i = 1, \dots, n$. A (directed) cycle is a (directed) path $(v_1, \dots, v_{n+1})$ such that $v_1 = v_{n+1}$. A path $(v_1, \dots, v_{n+1})$ is \emph{simple} if all vertices $v_1, \dots, v_{n+1}$ are distinct. We define an \emph{unshielded triple} in $G$ to be a tuple $(\{v,w\},u)$ for distinct vertices $v,w,u \in V$ such that $v,u$ and $w,u$ are adjacent but $v,w$ are not adjacent in $G$. If $v,w,u$ are all pairwise adjacent, then we say that $(\{v,w\},u)$ is a \emph{shielded triple}. We call an unshielded triple $(\{v,w\},u)$ in $G$ a \emph{collider} if $(v,u), (w,u) \in A(G)$ and a \emph{non-collider} otherwise. We say that a graph $G$ is \emph{undirected} or \emph{directed} if all of its edges are undirected or directed respectively. A \emph{(partially) directed acyclic graph (DAG/PDAG)} is a (partially) directed graph with no directed cycles. We define the set of parents of $w$ in $G$ as $\Pa(w, G) = \{v \mid (v,w) \in A(G)\}$, the set $\Desc(w, G)$ of \emph{descendants} of $w$ as the set of vertices $v$ such that there exists a directed path from $w$ to $v$ (this includes $w$), and the set of undirected neighbors of $w$ as $\Ne(w,G) = \{v \mid \{v,w\} \in E(G)\}$. For an undirected graph $G$, we define $\Ne^m(v, G)$ to be the set of vertices $w$ for which there exists a path from $w$ to $v$ of length at most $m$. For a DAG $G$, distinct vertices $v,w$ and a vertex set $Z \subseteq V(G)$ with $Z \cap \{v,w\} = \emptyset$, we say that a simple path $(v=v_1, v_2, \dots, v_n = w)$ is d-connecting $v$ and $w$ given $Z$ if for all $i = 2, \dots, n-1$, we have $v_i \notin Z$ if the triple $(\{v_{i-1}, v_{i+1}\}, v_i)$ is a non-collider and $\Desc(v_i) \cap Z \neq \emptyset$ otherwise. If a (simple) d-connecting path between $v$ and $w$ exists in $G$ given $Z$, then we write $(v \not \perp w \mid Z)_G$ and otherwise, we write $(v \perp w \mid Z)_G$ and say that $v$ and $w$ are d-separated given $Z$ in $G$. We say that two DAGs $G_1$ and $G_2$ are \emph{Markov-equivalent} if for all $v,w,Z$, we have $(v \perp w \mid Z)_{G_1} \Leftrightarrow (v \perp w \mid Z)_{G_2}$. We denote the \emph{Markov equivalence class} of a DAG $G$, that is the set of all DAGs that are Markov-equivalent to $G$, by $[G]$. A \emph{completed partially directed acyclic graph (CPDAG)} is a PDAG that graphically represents a Markov-equivalence class $[G]$ as follows: it has a directed edge $v \to w$ if and only if all DAGs $G' \in [G]$ contain the edge $v \to w$ and it has an undirected edge $\{v,w\}$ if there exist $G_1, G_2 \in [G]$ such that $v \to w$ in $G_1$ and $w \to v$ in $G_2$. Let $\C$ be the CPDAG representing the Markov equivalence class $[G]$. Since all DAGs in $[G]$ have the same d-connection and d-separation relationships, we extend our notation and write $(v \perp w \mid Z)_{\C}$ if $v$ and $w$ are d-separated in all graphs $G' \in [G]$ and $(v \not \perp w \mid Z)_{\C}$ if $v$ and $w$ are d-connected in all graphs $G' \in [G]$. We define a \emph{consistent DAG extension} of a PDAG $\C$ to be a DAG $G$ with the same adjacencies and unshielded colliders as $\C$ such that each directed edge in $\C$ is also a directed edge in $G$. \citet{verma_equivalence_1990} proved that the consistent DAG extensions of a CPDAG $\C$ are exactly the DAGs in the Markov-equivalence class represented by $\C$.

\textbf{Markov and faithfulness conditions.} The following proposition connects the concept of d-separation to conditional independence for linear Gaussian SEMs:

\begin{prop}[Markov condition for linear SEMs~\citep{pearl_causality_2009}] \label{prop:CMC}
    Let $(A, \bv)$ be a linear Gaussian SEM for $\bX = (X_1, \dots, X_n)$ and corresponding DAG $G$. Then, for all $i,j \in [n]$ and $Z \subseteq [n] \setminus \{i,j\}$, we have $(i \perp j \mid Z)_G \implies X_i \perp X_j \mid X_Z$.
\end{prop}

The converse of the statement above is not guaranteed to hold. Instead, it is often assumed as the standard faithfulness condition:

\begin{defn}[faithfulness] \label{defn:standard-faithfulness}
    We say that a linear Gaussian SEM $(A, \bv)$ for $\bX = (X_1, \dots, X_n)$ with corresponding DAG $G$ satisfies the \emph{faithfulness condition} if for all $i,j \in [n]$ and $Z \subseteq [n] \setminus \{i,j\}$, we have $(i \not \perp j \mid Z)_G \implies X_i \not \perp X_j \mid X_Z$.
\end{defn}

Multiple weaker versions of the faithfulness condition have been studied as well:

\begin{defn}[Triangle faithfulness \citep{zhang_detection_2008}] \label{defn:triangle-faithfulness}
     A linear Gaussian SEM for $\bX = (X_1, \dots, X_n)$ with corresponding DAG $G$ satisfies \emph{triangle faithfulness} if for any shielded non-collider triple $(\{i,j\}, k)$ in $G$, we have $X_i \not \perp X_j \mid X_Z$ for all $Z \subseteq [n] \setminus \{i,j,k\}$, and for any shielded collider triple $(\{i,j\}, k)$, we have $X_i \not \perp X_j \mid X_Z \cup \{X_k\}$ for all $Z \subseteq [n] \setminus \{i,j,k\}$.
\end{defn}

\begin{defn}[Adjacency faithfulness \citep{ramsey_adjacency-faithfulness_2006}] \label{defn:adjacency-faithfulness}
    A linear Gaussian SEM for $\bX = (X_1, \dots, X_n)$ with corresponding DAG $G$ satisfies \emph{adjacency faithfulness} if for any $i,j$ that are adjacent in $G$ and $Z \subseteq [n] \setminus \{i,j\}$, we have $X_i \not \perp X_j \mid X_Z$.
\end{defn}

\textbf{Useful mathematical statements.}
We state some known results on graphical models that we will employ in our proofs. The following is a simple observation about graphical d-separation:

\begin{lemma}[\citep{pearl_causality_2009}] \label{lem:parentseparation}
    Let $G$ be a DAG and let $v,w \in V(G)$ such that $v \notin \Pa(w) \cup \Desc(w)$. Then, $v$ and $w$ are d-separated in $G$ given $\Pa(w)$. 
\end{lemma}

Moreover, we will need a graphical characterization of when there exists a consistent DAG extension for certain partially directed acyclic graphs. 

 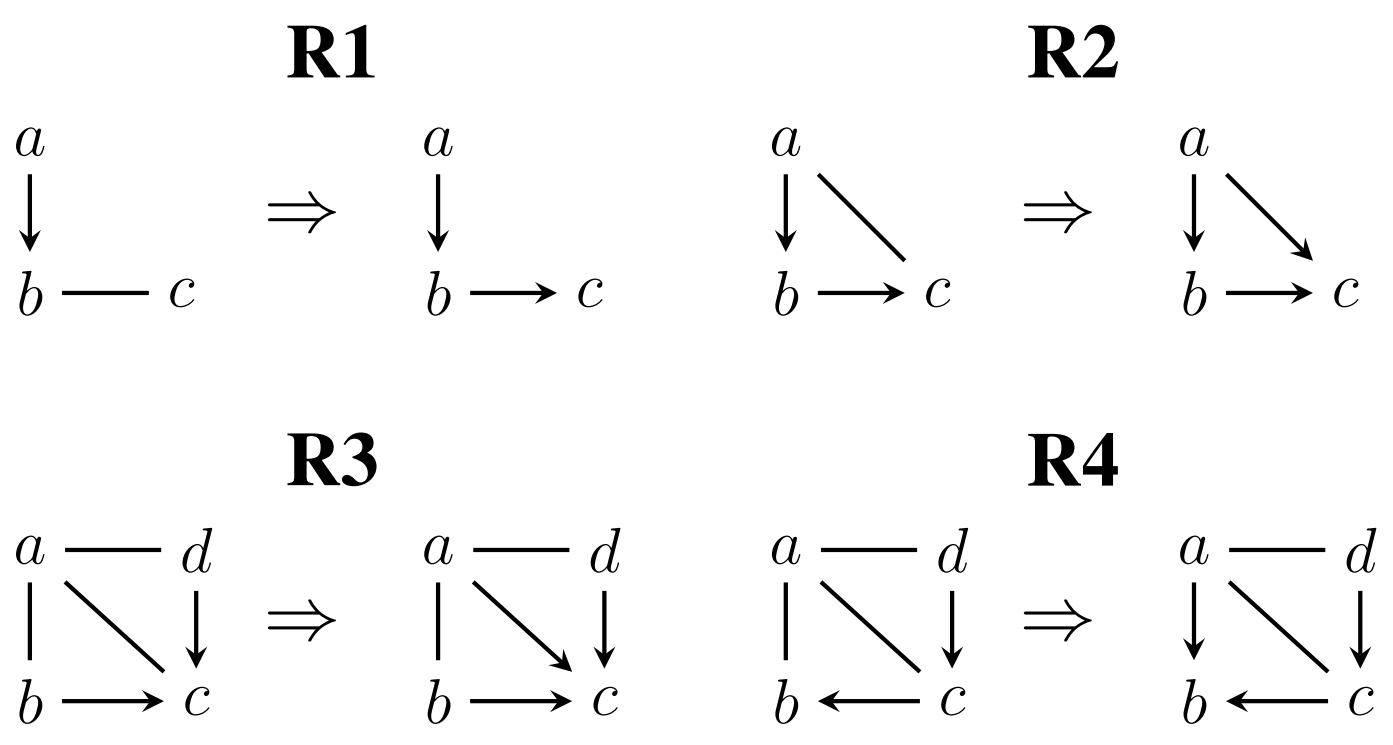
\begin{figure}
    \centering
    \begin{tikzpicture}[
    node distance=0.9cm,
    every node/.style={font=\large},
    arr/.style={->, thick, >=stealth},
    line/.style={thick},
    implies/.style={font=\Large}
]

\node[font=\bfseries\Large] at (2.0,2.2) {R1};

\node (r1a1) at (0,1.6) {$a$};
\node (r1b1) at (0,0.6) {$b$};
\node (r1c1) at (1.0,0.6) {$c$};
\draw[arr] (r1a1) -- (r1b1);
\draw[line] (r1b1) -- (r1c1);

\node[implies] at (1.8,1.1) {$\Rightarrow$};

\node (r1a2) at (2.7,1.6) {$a$};
\node (r1b2) at (2.7,0.6) {$b$};
\node (r1c2) at (3.7,0.6) {$c$};
\draw[arr] (r1a2) -- (r1b2);
\draw[arr] (r1b2) -- (r1c2);

\node[font=\bfseries\Large] at (6.9,2.2) {R2};

\node (r2a1) at (5.0,1.6) {$a$};
\node (r2b1) at (5.0,0.6) {$b$};
\node (r2c1) at (6.0,0.6) {$c$};
\draw[arr] (r2a1) -- (r2b1);
\draw[arr] (r2b1) -- (r2c1);
\draw[line] (r2a1) -- (r2c1);

\node[implies] at (6.8,1.1) {$\Rightarrow$};

\node (r2a2) at (7.7,1.6) {$a$};
\node (r2b2) at (7.7,0.6) {$b$};
\node (r2c2) at (8.7,0.6) {$c$};
\draw[arr] (r2a2) -- (r2b2);
\draw[arr] (r2b2) -- (r2c2);
\draw[arr] (r2a2) -- (r2c2);

\node[font=\bfseries\Large] at (2.0,-0.5) {R3};

\node (r3a1) at (0,-1.1) {$a$};
\node (r3b1) at (0,-2.1) {$b$};
\node (r3c1) at (1.1,-2.1) {$c$};
\node (r3d1) at (1.1,-1.1) {$d$};
\draw[line] (r3a1) -- (r3b1);
\draw[line] (r3a1) -- (r3d1);
\draw[line] (r3a1) -- (r3c1);
\draw[arr] (r3b1) -- (r3c1);
\draw[arr] (r3d1) -- (r3c1);

\node[implies] at (1.8,-1.6) {$\Rightarrow$};

\node (r3a2) at (2.7,-1.1) {$a$};
\node (r3b2) at (2.7,-2.1) {$b$};
\node (r3c2) at (3.8,-2.1) {$c$};
\node (r3d2) at (3.8,-1.1) {$d$};
\draw[line] (r3a2) -- (r3b2);
\draw[line] (r3a2) -- (r3d2);
\draw[arr] (r3a2) -- (r3c2);
\draw[arr] (r3b2) -- (r3c2);
\draw[arr] (r3d2) -- (r3c2);

\node[font=\bfseries\Large] at (6.9,-0.5) {R4};

\node (r4a1) at (5.0,-1.1) {$a$};
\node (r4b1) at (5.0,-2.1) {$b$};
\node (r4c1) at (6.1,-2.1) {$c$};
\node (r4d1) at (6.1,-1.1) {$d$};
\draw[line] (r4a1) -- (r4b1);
\draw[line] (r4a1) -- (r4d1);
\draw[line] (r4a1) -- (r4c1);
\draw[arr] (r4c1) -- (r4b1);
\draw[arr] (r4d1) -- (r4c1);

\node[implies] at (6.8,-1.6) {$\Rightarrow$};

\node (r4a2) at (7.7,-1.1) {$a$};
\node (r4b2) at (7.7,-2.1) {$b$};
\node (r4c2) at (8.8,-2.1) {$c$};
\node (r4d2) at (8.8,-1.1) {$d$};
\draw[arr] (r4a2) -- (r4b2);
\draw[line] (r4a2) -- (r4d2);
\draw[line] (r4a2) -- (r4c2);
\draw[arr] (r4c2) -- (r4b2);
\draw[arr] (r4d2) -- (r4c2);

\end{tikzpicture}

    \caption{The four Meek rules: the uncompleted configurations are on the left-hand side for each rule and in each case, they are completed by orienting one additional edge.}
    \label{fig:meek}
\end{figure}

\begin{defn}
    A \emph{maximally oriented partially directed acyclic graph (MPDAG)} is a PDAG that contains none of the uncompleted Meek configurations in Figure~\ref{fig:meek}. 
\end{defn}

An \emph{undirected component} of a partially directed graph $G$ is a maximal connected undirected subgraph of $G$. Each partially directed graph is partitioned by its undirected components. For a subgraph $S \subseteq G$, we define the \emph{induced graph} $G[S]$ as the subgraph of $G$ that consists of the vertex set $V(S)$ and all edges of $G$ between vertices in $V(S)$. 

\begin{theorem}[\citet{wienobst_extendability_2021}] \label{thm:ext-mpdag}
    Let $G$ be an MPDAG with undirected components $S_1, \dots, S_m$. Then $G$ has a consistent DAG extension if and only if none of the graphs $G[S_i]$ for $i=1, \dots, m$ contain an induced cycle of length at least $4$. Moreover, running Algorithm~\ref{alg:wienobst} on each $G[S_i]$ produces a consistent DAG extension if it exists. 
\end{theorem}

\textbf{Partial orderings.}
For context, we provide some definitions related to partially ordered sets, underlying the framework of~\citet{TBC24} for false positive error. 

\begin{defn}[Partial order]
Let $P$ be a set. A binary relation $\preceq$ on $P$ is called a
\emph{partial order} if, for all $x,y,z \in P$, the following hold:
\begin{enumerate}
    \item Reflexivity: $x \preceq x$.
    \item Antisymmetry: if $x \preceq y$ and $y \preceq x$, then $x=y$.
    \item Transitivity: if $x \preceq y$ and $y \preceq z$, then $x \preceq z$.
\end{enumerate}
A set equipped with a partial order is called a \emph{partially ordered set}
or \emph{poset}, and is denoted by $(P,\preceq)$.
\end{defn}

\begin{defn}[Graded partial order]
Let $(P,\preceq)$ be a partially ordered set. We say that $(P,\preceq)$ is
\emph{graded} if there exists a function $r: P \to \mathbb{Z}_{\geq 0}$,
called a \emph{rank function}, such that whenever $y$ covers $x$, meaning
\begin{align*}
    x \prec y
    \quad\text{and there is no } z \in P \text{ with } x \prec z \prec y,
\end{align*}
we have
\begin{align*}
    r(y) = r(x) + 1.
\end{align*}
\end{defn}

The rank function defines a clear hierarchy for the elements of a graded poset and rank difference is a useful notion for distance in graded posets, which motivates Definition~\ref{defn:FD}. It is a non-trivial theorem by~\citet{Chickering03} that the number of adjacencies of a CPDAG is a valid rank function for the poset that we define in Definition~\ref{defn:cpdag-poset}.

\textbf{PC algorithm and variants.}
Since we use in-depth comparisons between our algorithm, the standard \emph{PC algorithm} and the \emph{conservative PC algorithm}, we state both these algorithms for completeness below:

\begin{algorithm} 
    \caption{\textsc{PC Algorithm} \citep{spirtes_causation_2001}} \label{alg:pc}
     \begin{lstlisting}[style=pseudocode,mathescape,numbers=left, firstnumber=0, escapechar=|,columns=fullflexible,breaklines=true]
     Input: set of samples $\cS$ of $\bX = (X_1,\dots,X_n)$, significance level $\alpha$.   
    $\widehat{G} \leftarrow$ complete undirected graph on $[n]$
    $\Ihat \leftarrow \emptyset$
    $\ell \leftarrow 0$ 
    while there exists an edge $\{i,j\}$ in $\widehat{G}$ with $|\Ne(i,\widehat{G}) \setminus \{j\}| \geq \ell$:
        for all edges $\{i,j\}$ in $\widehat{G}$:
            for all $Z \subseteq \Ne(i,\widehat{G}) \setminus \{j\}$ with $|Z|=\ell$:
                if $X_i \perp X_j \mid X_Z$ is accepted by |\textsc{ConditionalIndependenceTest($\cS,i,j,Z,\alpha$)}|:
                    delete $\{i,j\}$ from $\widehat{G}$
                    $\Ihat \leftarrow \Ihat \cup \{(\{i,j\},Z)\}$
                    break
        $\ell \leftarrow \ell+1$

    for all unshielded triples $i-k-j$ in $\widehat{G}$:
        if $k \notin Z$ for all $(\{i,j\},Z) \in \Ihat$:
            orient $i-k-j$ as $i \to k \leftarrow j$
    Keep applying Meek rules to $\widehat{G}$ |(see Figure~\ref{fig:meek})| that only involve unambiguous triples, until none of them is applicable 
    return $\widehat{G}$
     \end{lstlisting}  
 \end{algorithm}

 \begin{algorithm} 
    \caption{\textsc{conservative PC Algorithm} \citep{ramsey_adjacency-faithfulness_2006}} \label{alg:cpc}
     \begin{lstlisting}[style=pseudocode,mathescape,numbers=left, firstnumber=0, escapechar=|,columns=fullflexible,breaklines=true]
     Input: set of samples $\cS$ of $\bX = (X_1,\dots,X_n)$, significance level $\alpha$.   
    $\Ehat \leftarrow$ complete undirected graph on $[n]$
    $\Ihat \leftarrow \emptyset$
    $\ell \leftarrow 0$ 
    while there exists an edge $\{i,j\}$ in $\Ehat$ with $|\Ne(i,\Ehat) \setminus \{j\}| \geq \ell$:
        for all edges $\{i,j\}$ in $\Ehat$:
            for all $Z \subseteq \Ne(i,\Ehat) \setminus \{j\}$ with $|Z|=\ell$:
                if $X_i \perp X_j \mid X_Z$ is accepted by |\textsc{ConditionalIndependenceTest($\cS,i,j,Z,\alpha$)}|:
                    delete $\{i,j\}$ from $\Ehat$
                    $\Ihat \leftarrow \Ihat \cup \{(\{i,j\},Z)\}$
                    break
        $\ell \leftarrow \ell+1$

    for all unshielded triples $i-k-j$ in $\Ehat$:
        for all $Z \subseteq \Ne(i,\Ehat)$ and all $Z \subseteq \Ne(j,\Ehat)$:
            if $X_i \perp X_j \mid X_Z$ is accepted by |\textsc{ConditionalIndependenceTest($\cS,i,j,Z,\alpha$)}|:
                $\Ihat \leftarrow \Ihat \cup \{(\{i,j\},Z)\}$
        if $k \notin Z$ for all $(\{i,j\},Z) \in \Ihat$:
            orient $i-k-j$ as $i \to k \leftarrow j$ in $\Ehat$
        else if $k \in Z$ for all $(\{i,j\},Z) \in \Ihat$:
            continue
        else: 
            mark $i-k-j$ as ambiguous in $\Ehat$
    Keep applying Meek rules to $\Ehat$ |(see Figure~\ref{fig:meek})| that only involve unambiguous triples, until none of them is applicable 
    return $\Ehat$
     \end{lstlisting}  
 \end{algorithm}

\newpage

\section{Algorithms and proofs} \label{apx:proofs}

We begin by proving the statements from section~\ref{sec:faithfulness} on $m$-connected collider faithfulness:

\faithfultriangles*

\begin{proof}[Proof]
By equation~\eqref{eq:sigmas}, we have 
\begin{align*}
    \Sigma_\bX^{-1} = (I - A)^{\top} \diag(\bv)^{-1} (I-A).
\end{align*}
In particular, for $i < j$, we get
\begin{align*}
    (\Sigma_\bX^{-1})_{ij} = \sum_{k \in C(i,j)} A_{ki}v_k^{-1}A_{kj} - A_{ji}v_j^{-1},
\end{align*}
because $A$ is lower-triangular and $k \in C(i,j)$ if and only if $A_{ki}A_{kj} \neq 0$.
Recall that $\{i,j\}$ is an edge in $H$ iff $(\Sigma_\bX^{-1})_{ij}$ is nonzero and by the 
equation above and Definition~\ref{defn:faithful_colliders}, this is iff the set of colliders
based at $(i,j)$ is faithful. 
\end{proof}

\distance*

\begin{proof}[Proof]
Denote $G=G(A)$, $H=H(A,\bv)$ and consider an edge $i \to j$ in $G$. If the set of colliders based at $(i,j)$ is faithful, then $\{i,j\}\in E(H)$ by Lemma~\ref{lem:faithful_triangles}, and we are done.
Hence, it remains to consider the case that the set of colliders based at $(i,j)$ is unfaithful.

For every edge $x \to y$ in $G$, such that the set of colliders based at $(x,y)$ is unfaithful, fix 
an arbitrary collider vertex $c(x,y)\in C(x,y)$. This is possible because unfaithfulness of the colliders implies
\begin{align*}
    \sum_{k\in C(x,y)} A_{kx}A_{ky}v_k^{-1}=A_{yx}v_y^{-1}\neq 0,
\end{align*}
and therefore $C(x,y)\neq \emptyset$.
We now iteratively construct a set $U$ of edges with unfaithful sets of colliders as follows: 
\begin{itemize}
    \item start with $U_0 = \{i \to j\}$;
    \item given $U_t$ at step $t$, consider the set of edges $N = \{x \to c(x,y), y \to c(x,y) \mid x \to y \in U_t\}$,
    and obtain $U_{t+1}$ by adding all edges of $N$ to $U_t$ at which the corresponding set of colliders is unfaithful. 
\end{itemize}

Since $G$ is finite, this process eventually stops with a set $U$ of edges such that the colliders based
at each edge are unfaithful. We show by induction that the subgraph of $G$ given by each $U_t$ has the unique root $i$. 
Clearly, this is true for $U_0$. Assuming it is true for $U_t$, consider a vertex $z$ in $U_{t+1}$ that is not in $U_t$.
We must have $z = c(x,y)$ for some vertices $x,y$ in $U_t$ and either $x \to z$ or $y \to z$ is in $U_{t+1}$. By the induction hypothesis, there 
are paths from $i$ to $x$ and $y$ in $U_t$, and one of them then extends to $z$ in $U_{t+1}$. Hence, there is a path from $i$
to every vertex in $U_{t+1}$ and since the subgraph given by $U_{t+1}$ is a DAG, the vertex $i$ must be a unique root. 

Now, we can invoke the $m$-connected-collider faithfulness condition to deduce that $U$ has at most $m-1$ edges. Next, 
consider the vertex set 

\begin{align*}
V^* = \{i,j\} \cup \{c(x,y) \mid x \to y \in U\}.
\end{align*}\

We get that $|V^*| \leq 2 + |U| \leq m+1$. The final step of the proof is to show that for each edge $x\to y \in U$, 
the endpoints $x,y$ are connected in the induced graph $H[V^*]$. Consider a consistent vertex ordering of $G$ (that is, an ordering $\prec$ on $V(G)$ such that $v \prec w$ implies $w \notin \Desc(v)$) and extend
it lexicographically to the edges of $G$. Then, consider the minimal edge $x \to y \in U$ such that $x,y$ are disconnected
in $H[V^*]$. Let $z = c(x,y) \in V^*$. By construction and Lemma~\ref{lem:faithful_triangles}, the edges $x \to z$ and $y \to z$ are either also undirected edges 
in $H[V^*]$ or they are edges in $U$ that are strictly smaller than $x \to y$. In both cases, $x,z$ and $y,z$ are connected 
in $H[V^*]$. But this implies, $x,y$ are connected in $H[V^*]$, a contradiction. 

We conclude that $i,j$ are connected in $H[V^*]$ and since $|V^*| \leq m+1$, we must have $\dist_H(i,j) \leq m$.

\end{proof}

Next, we formally state our multiple testing algorithm and prove its error guarantee as stated in Section~\ref{sec:multiple-testing}.

\begin{algorithm} 
    \caption{\textsc{TestConditionalIndependence}} \label{alg:fdr}
     \begin{lstlisting}[style=pseudocode,mathescape,numbers=left, firstnumber=0, escapechar=|,columns=fullflexible,breaklines=true]
     Input: set $\cL$ of triples $(\{i,j\},Z)$ and $p$-values $p_{i,j,Z}$ corresponding to the null hypotheses $\cH_{0,i,j,Z}: X_i \perp X_j \mid X_Z$, significance level $\alpha$
    $M \leftarrow \{\{i,j\} \mid \exists Z: (\{i,j\},Z) \in \cL\}$, $m \leftarrow |M|$
    for all $i < j$ with $\{i,j\} \in M$:
        $P_{i,j} \leftarrow \max_{Z:(\{i,j\},Z) \in \cL} p_{i,j,Z}$
    sort the $P_{i,j}$ in increasing order, denote the result $P_{(1)} \leq \dots \leq P_{(m)}$
    $K \leftarrow \max \{k : P_{(k)} \leq k \alpha / (m H_m) \}$ where $H_m = \sum_{r=1}^m 1/r$
    $\Ihat \leftarrow \emptyset$
    for all $(\{i,j\},Z) \in \cL$:
        if $p_{i,j,Z} > K \alpha / (m H_m)$:
            $\Ihat \leftarrow \Ihat \cup \{(\{i,j\},Z)\}$
    return $\Ihat$
    
     \end{lstlisting}  
 \end{algorithm}

\fdrcontrol*

\begin{proof}[Proof]
    Our proof idea is derived from \citet{wang_elementary_2025}. Suppose that for all $(\{i,j\}, Z) \in \cL$, the variables $p_{i,j,Z}$ are valid $p$-values for the null-hypotheses $\cH_{0,i,j,Z}: X_i \perp X_j \mid X_Z$ and define $m, P_{i,j}, P_{(k)}, K, H_m$ as in Algorithm~\ref{alg:fdr}. Let $N = \{(i,j) \mid i<j, i \notin \Pa(j, G^*)\}$ and denote $p^*_{i,j} = p_{i,j,\Pa(j, G^*)}$ for $(i,j) \in N$, and $k\alpha / (mH_m) = \alpha_k$. Note that by assumption, we have $(\{i,j\}, \Pa(j)) \in \cL$ for all $(i,j) \in N$. Moreover, the pairs $\{i,j\}$ for which $p_{i,j,Z}$ gets rejected for all $Z$ with $(\{i,j\}, Z) \in \cL$ are exactly the pairs corresponding to $P_{(1)}, \dots, P_{(K)}$. Hence, we have $\Ad(\Ihat) = K$. We write
    \begin{align*}
        \E\left[\frac{\FPP(\Ihat, G^*)}{\Ad(\Ihat)}\right] 
        &= \E\left[\sum_{(i,j)\in N}\frac{\one_{\{p^*_{i,j} \leq \alpha_K\}}}{K}\right]\\
        &= \sum_{(i,j)\in N} \sum_{k=1}^{m}\frac{1}{k}\E\left[\one_{\{p^*_{i,j} \leq \alpha_K\}}\one_{\{K = k\}}\right]
    \end{align*}
    Using the telescoping identity 
    \begin{align*}
        \frac{1}{k} = \sum_{s = k}^{m} \left(\frac{1}{s} - \frac{1}{s+1}\right) + \frac{1}{m + 1} = \sum_{s = k}^{m} \frac{1}{s(s+1)} + \frac{1}{m + 1},
    \end{align*}
    we get
    \begin{align*}
        \E\left[\frac{\FPP(\Ihat, G^*)}{\Ad(\Ihat)}\right] 
        &= \sum_{(i,j) \in N} \sum_{k=1}^{m}\left(\sum_{s = k}^{m} \frac{1}{s(s+1)} + \frac{1}{m + 1}\right)\E\left[\one_{\{p^*_{i,j} \leq \alpha_K\}}\one_{\{K = k\}}\right]\\
        &= \sum_{(i,j) \in N} \sum_{s=1}^{m}\frac{1}{s(s+1)}\sum_{k = 1}^{s} \E\left[\one_{\{p^*_{i,j} \leq \alpha_K\}}\one_{\{K = k\}}\right]\\ 
        &+  \frac{1}{m + 1}\sum_{(i,j) \in N}\sum_{k = 1}^{m}\E\left[\one_{\{p^*_{i,j} \leq \alpha_K\}}\one_{\{K = k\}}\right]\\
        &= \sum_{(i,j) \in N} \sum_{s=1}^{m}\frac{1}{s(s+1)}\E\left[\one_{\{p^*_{i,j} \leq \alpha_K\}}\one_{\{K \leq s\}}\right] +  \frac{1}{m + 1}\sum_{(i,j) \in N}\E\left[\one_{\{p^*_{i,j} \leq \alpha_K\}}\right]\\
    \end{align*}
    Now, using the facts that $p^*_{i,j}$ is a valid $p$-value for all $i,j$, $\alpha_k$ is increasing in $k$, and $|N| \leq m$, we get 
    \begin{align*}
        \E\left[\frac{\FPP(\Ihat, G^*)}{\Ad(\Ihat)}\right] 
        &\leq \sum_{(i,j) \in N} \left(\sum_{s=1}^{m}\frac{\alpha_s}{s(s+1)} +  \frac{\alpha_{m}}{m + 1}\right)\\
        &= \frac{|N|\alpha}{mH_{m}}\left(\sum_{s=1}^{m}\frac{1}{(s+1)} +  \frac{m}{m + 1}\right)\\
        &=\frac{|N|\alpha}{m} \leq \alpha.\\
    \end{align*}
\end{proof}

Next, we give detailed pseudocode for all the steps in our orientation procedure:

\begin{algorithm} 
    \caption{\textsc{OrientEPattern}} \label{alg:orient}
     \begin{lstlisting}[style=pseudocode,mathescape,numbers=left, firstnumber=0, escapechar=|,columns=fullflexible,breaklines=true]
     Input: set $\cI$ of tuples $(\{i,j\},Z)$ indicating conditional independencies and undirected graph $\cE$
    $T \leftarrow \{(\{i,j\},k) \mid \{i,k\}, \{j,k\} \in E(\cE), \{i,j\} \notin E(\cE) \}$ (set of unshielded triples)
    $A \leftarrow \emptyset, C \leftarrow \emptyset, NC \leftarrow \emptyset$
    for $(\{i,j\},k) \in T$:
        if $\exists Z$ with $k \notin Z$ and $(\{i,j\},Z) \in \cI$:
            if $\exists Y$ with $k \in Y$ and $(\{i,j\},Y) \in \cI$:
                $A \leftarrow A \cup \{(\{i,j\},k)\}$
            else:
                $C \leftarrow C \cup \{(\{i,j\},k)\}$
        else: 
            $NC \leftarrow NC \cup \{(\{i,j\},k)\}$
    Define $\cF(A, C, NC) := \{f:T \to \{c, nc\} \mid \forall T \in C: f(T) = c, \forall T \in NC: f(T) = nc \}$ 
    $\cF \leftarrow \cF(A, C, NC)$ 
    $\cG \leftarrow \emptyset$, $\cF^* \leftarrow \emptyset $
    while $\cF \neq \emptyset$ or $\cG \neq \emptyset$:
        if $\cF \neq \emptyset:$
            take an assignment $f \in \cF$
            $\cF^* \leftarrow \cF^* \cup \{f\}$
            if |\textsc{RealizeAssignment}($\cE, f$)| outputs a valid DAG $G$: |\label{line-realize-assignment}|
                $\cG \leftarrow \cG \cup \{G\}$
                $\cF \leftarrow \cF \setminus \{f\}$
            else:
                let $R$ be the set of conflicting triples in the output of |\textsc{RealizeAssignment}($\cE, f$)|
                $A \leftarrow A \cup R$, $C \leftarrow C \setminus R$, $NC \leftarrow NC \setminus R$ |\label{line:create-ambiguities}|
                $\cF \leftarrow \cF(A, C, NC) \setminus \cF^*$
        else:
            take a DAG $G$ from $\cG$
            $\cG \leftarrow \cG \setminus \{G\}$
            for all $(\{i,j\},Z) \in \cI$ such that $(i \not \perp j \mid Z)_G$:
                 for all simple paths $\pi$ in $G$ that d-connect $i,j$ given $Z$:
                    let $W(\pi)$ be the union of $\pi$ together with all directed paths from any collider on $\pi$ to $Z$
                    let $R \subseteq T$ be the set of unshielded triples whose edges are in $W(\pi)$
                    $A \leftarrow A \cup R$, $C \leftarrow C \setminus R$, $NC \leftarrow NC \setminus R$ |\label{line:create-ambiguities-2}|
                    $\cF \leftarrow \cF(A, C, NC) \setminus \cF^*$
    for all $T \in A$:
        mark $T$ as ambiguous in $\cE$
    for all $T= (\{i,j\},k) \in C$:
        orient $i \to k$ and $j \to k$ in $\cE$
    Keep applying Meek rules to $\cE$ |(see Figure~\ref{fig:meek})| that only involve non-ambiguous triples, until none of them is applicable 
    return $\cE$
     \end{lstlisting}  
 \end{algorithm}
 
 \begin{algorithm}
     \caption{\textsc{RealizeAssignment}} \label{alg:realize}
     \begin{lstlisting}[style=pseudocode,mathescape,numbers=left, firstnumber=0, escapechar=|,columns=fullflexible,breaklines=true]
     Input: undirected graph $G$ with unshielded triples $T$ and assignment $f: T \to \{c, nc\}$
     $C \leftarrow \{(\{i,j\},k) \in T \mid f(\{i,j\},k) = c\}, NC \leftarrow \{(\{i,j\},k) \in T \mid f(\{i,j\},k) = nc\}$
     for all $(\{i,j\},k) \in C$: |\label{start-pdag}|
        if $\exists a$ with $(\{a,k\},i) \in C$:
            return $R = \{(\{i,j\},k), (\{a,k\},i)\}$ 
        if $\exists b$ with $(\{b,k\},j) \in C$:
            return $R = \{(\{i,j\},k), (\{b, k\},j)\}$ 
    for all $(\{i,j\},k) \in NC$: |\label{start-ncoll-check}|
        if $\exists a,b$ with $(\{a,i\},k), (\{b,j\},k) \in C$:
            return $R = \{(\{i,j\},k), (\{a,i\},k), (\{b,j\},k)\}$ 
    initialize $T_e \leftarrow \emptyset$ for all $e \in E(G)$
    for all $t = (\{i,j\},k) \in C$:
        orient $i \to k$ and $j \to k$ in $G$
        $T_{\{i,k\}} \leftarrow T_{\{i,k\}} \cup \{t\}$, $T_{\{j,k\}} \leftarrow T_{\{j,k\}} \cup \{t\}$   
    if $G$ contains a directed cycle $H$: 
        return $R = \bigcup_{e \in H} T_e$ |\label{end-pdag}|
    while $G$ contains an uncompleted Meek configuration: |\label{start-mpdag}|
        let $e_1, \dots, e_r$ be the set of directed edges that complete any uncompleted Meek configuration in $G$ according to the rules R1 - R4 (see |Figure~\ref{fig:meek}|)
        for all $j = 1, \dots, r$:
            let $E_j$ be the set of directed edges in each uncompleted Meek configuration that forces $e_j$
            let $T_j$ be the set of unshielded triples in each uncompleted Meek configuration that forces $e_j$
            $T_{e_j} \leftarrow \bigcup_{e \in E_j} T_e \cup T_j$
        if $\exists 1\leq i,j \leq r$ such that  $e_i = v \to w, e_j = w \to v$:
            return $R = T_{e_i} \cup T_{e_j}$
        add the orientations of $e_1, \dots, e_r$ to $G$
        if $\exists T \in NC$ and two directed edges $e,f$ in $G$ that orient $T$ as a collider:
            return $R = \{T\} \cup T_e \cup T_f$
        if $G$ contains a directed cycle $H$: 
            return $R = \bigcup_{e \in H} T_e$  |\label{end-mpdag}|
    let $S_1, \dots, S_m$ be the undirected components of $G$ |\label{start-dag}|
    for all $i = 1, \dots, m$:
        if $G[S_i]$ contains an induced cycle $H = (v_1, \dots, v_r)$ of length at least $4$:
            return $R = \{(\{v_i, v_{i+2}\}, v_{i+1}) \mid i=1, \dots, r\}$ |\label{line-cycle}|
        else: 
            $G[S_i] \leftarrow $ |\textsc{OrientBucket}($G[S_i]$)| |\label{end-dag}|
    return $G$ |\label{line-final}|
    
     \end{lstlisting}  
 \end{algorithm}

  \begin{algorithm}
     \caption{\textsc{OrientBucket} \citep{wienobst_extendability_2021}} \label{alg:wienobst}
     \begin{lstlisting}[style=pseudocode,mathescape,numbers=left, firstnumber=0, escapechar=|,columns=fullflexible,breaklines=true]
     Input: PDAG $G$ 
     $S \leftarrow (V(G)), \tau \leftarrow ()$
     while $S = (S_1, S_2, \dots)$ is non-empty:
        choose a vertex $v$ with no incoming edges from $S$
        let $N(v)$ be the set of vertices adjacent to $v$
        $S \leftarrow (S_1 \cap N(v), S_1 \setminus N(v), S_2 \cap N(v), S_2 \setminus N(v), \dots)$
        $\tau \leftarrow \tau + v$
    orient the edges of $G$ according to the ordering $\tau$
    return $G$
    
     \end{lstlisting}  
 \end{algorithm}

\newpage 

Before we can prove our main statement (Proposition~\ref{prop:CIstatements}) about the orientation step from section~\ref{sec:orientation-step}, we need two auxiliary lemmata. Let $\pi$ be a d-connecting path from $v$ to $w$ given $Z$ in a DAG $G$. We define a \emph{witness graph} $W(\pi)$ for $\pi$ to be the union of $\pi$ together with shortest directed paths from each collider vertex on $\pi$ to a vertex in $Z$. 

\begin{lemma} \label{lem:d-connection}
    Let $G$ and $H$ be two DAGs on vertex set $V$ such that every adjacency in $H$ is also an adjacency in $G$. Consider vertices $v,w \in V$ and a vertex set $Z \subseteq V \setminus \{v,w\}$, such that $(v \not \perp w \mid Z)_H$ and suppose that for all d-connecting paths $\pi$ in $H$, there exists a witness graph $W(\pi)$ such that every unshielded triple in $W(\pi)$ has the same collider status as the corresponding triple in $G$. Then, $(v \not \perp w \mid Z)_G$. 
\end{lemma}

\begin{proof}
    Let $\pi$ be a path that d-connects $v,w$ given $Z$ in $H$ and that is minimal among all d-connecting paths with respect to $(|A(\pi)|, |A(W(\pi))|)$ lexicographically, where $W(\pi)$ is a witness graph that satisfies the condition in Lemma~\ref{lem:d-connection}. Since the adjacencies in $H$ are a subset of the adjacencies in $G$, the path $\pi$ also exists in $G$ and we will show that it is a d-connecting path. Let $(u,v,w)$ be a subpath of $\pi$. First, assume that the triple $(\{u,w\}, v)$ is a collider in $G$. If it is unshielded in $H$, then $(\{u,w\}, v)$ must also be a collider in $H$ by our assumptions. If $u,w$ were adjacent in $H$, then the path $\pi'$ that skips $v$ by using the edge between $u,w$ would also be a d-connecting path in $H$, which contradicts the minimality of $\pi$. Hence, every collider on $\pi$ in $G$ is also a collider in $H$. Next, consider a triple $(\{u,w\}, v)$ that is a non-collider in $G$. Again, if it is unshielded in $H$, then $(\{u,w\}, v)$ must also be a non-collider in $H$ by our assumptions. If $u,w$ are adjacent in $H$, then by minimality of $\pi$, the path $\pi'$ that skips $v$ by using the edge between $u,w$ cannot be d-connecting in $H$. This is only possible if $u \leftarrow v$ and $v \to w$ in $H$ and either $u$ or $w$ is a collider on $\pi$ in $H$. But then again, $(\{u,w\}, v)$ must be a non-collider in $H$. Since $\pi$ is a d-connecting path in $H$, this implies that $v \notin Z$ for all non-colliders on $\pi$ in $G$. It remains to show that for any subpath $(u,v,w)$ of $\pi$ where $v$ is a collider in $G$, we have $\Desc(v, G) \cap Z \neq \emptyset$. We already showed that $v$ must be a collider on $\pi$ in $H$, so let $\tau = (x_1 = v, x_2, \dots, x_t \in Z)$ be the directed path in $H$ from $v$ to $Z$ that is part of $W(\pi)$. None of the triples $(\{x_i, x_{i+2}\}, x_{i+1})$ can be shielded in $H$, because otherwise we would get a directed path from $v$ to $Z$ that skips $x_{i+1}$ contradicting that $\tau$ is minimal. Hence, by our assumptions, all these triples must also be non-colliders in $G$. This means that $\tau$ is either a directed path from $v$ to $Z$ in $G$ or we have $v \leftarrow x_2$ in $G$. In that case, the triples $(\{u, x_2\}, v)$ and $(\{w, x_2\}, v)$ must be shielded in $H$, because they have a different collider status in $G$, but then, replacing $(u,v,w)$ in $\pi$ by $(u,x_2,w)$ leads to a path $\pi'$ in $H$ with $|A(\pi')| = |A(\pi)|$ and $|A(W(\pi'))| < |A(W(\pi))|$ contradicting minimality. Hence, $\tau$ must indeed be a directed path from $v$ to $Z$ in $G$, which means that $\pi$ is d-connecting in $G$. 
\end{proof}

\begin{defn}
    Let $S$ be an undirected graph and let $T = \{(\{i,j\},k) \mid \{i,k\}, \{j,k\} \in E(S), \{i,j\} \notin E(S)\}$ be the set of unshielded triples of $S$. Let $f: T \to \{c, nc\}$ be an assignment of each unshielded triple to either ``collider'' ($c$) or ``non-collider'' ($nc$). We say that $f$ is \emph{realizable} on $S$ if there exists an orientation of the edges of $S$ such that the resulting graph $G$ is a DAG and all its unshielded colliders and non-colliders agree with the assignment $f$. In this case, we also say that $G$ \emph{realizes} the assignment $f$ on $S$. 
\end{defn}

\begin{lemma} \label{lem:realize}
    Let $S$ be an undirected graph, let $T$ be the set of unshielded triples of $S$ and let $f:T \to \{c, nc\}$ be an assignment of each triple to either a collider or a non-collider. If the output of $\textsc{RealizeAssignment}(S, f)$ (Algorithm~\ref{alg:realize}) is a graph $G$, then it is a valid DAG that realizes $f$ on $S$. Otherwise, if Algorithm~\ref{alg:realize} outputs a set $R \subseteq T$ instead of a DAG, then there exists no DAG that realizes any assignment $g$ on $S$ that agrees with $f$ on $R$. 
\end{lemma}

\begin{proof}
   If the algorithm outputs a set of triples $R$ before line~\ref{end-pdag}, then these triples either have contradicting collider/non-collider assignments or collider assignments that force a directed cycle. Hence, no DAG on $S$ that is consistent with the assignment $f$ on $R$ can exist. Once the algorithm passes line~\ref{end-pdag}, the graph $G$ must then be a valid PDAG that is consistent with the assignment $f$. In lines~\ref{start-mpdag} to~\ref{end-mpdag}, the algorithm outputs a set of triples $R$ if and only if the collider/non-collider assignment on $R$ forces a contradiction or a directed cycle according to the Meek rules. By the soundness of the Meek rules~\citep{meek_causal_1995}, no DAG on $S$ that is consistent with the collider assignment on $R$ can exist in this case. Once the algorithm passes line~\ref{end-mpdag}, the graph $G$ must be a valid MPDAG that is consistent with $f$. If the algorithm outputs a set $R$ in line~\ref{line-cycle}, then, by Lemma 5.2 in~\citet{wienobst_extendability_2021}, all triples in $R$ must have been assigned non-collider status by $f$. However, no DAG can exist that has an induced cycle of length at least $4$, on which each unshielded triple is a non-collider (since any consistent orientation of such a cycle must create a directed cycle). Otherwise, if the algorithm passes to line~\ref{line-final}, then the output must be a valid DAG that realizes $f$ on $S$ by Theorem~\ref{thm:ext-mpdag}.
\end{proof}

Now, we are ready to prove the theoretical property that Algorithm~\ref{alg:orient} is designed for. 

\orientation*

\begin{proof}[Proof]
    Let $T$ be the set of unshielded triples of $S$. Let $\widetilde{G}^*$ be the DAG that is obtained from adding all adjacencies of $S$ that are not in $G^*$ to $G^*$ and orienting the edges according to a consistent vertex ordering of $G^*$. Now, consider the assignment $f: T \to \{c,nc\}$ that assigns $c$ exactly to all unambiguous colliders of $\Ehat$ and to all ambiguous unshielded triples of $\Ehat$ that are (shielded or unshielded) colliders in $\widetilde{G}^*$. First, we prove that $f$ is realizable by a DAG $\widehat{G}$ on $S$. Since $f$ is an assignment of triples that is consistent with the unambiguous unshielded triples in $\Ehat$, it must have been added to $\cF$ at some point in Algorithm~\ref{alg:orient} and therefore Algorithm~\ref{alg:orient} must have run $\textsc{RealizeAssignment}(S, f)$. Suppose that $\textsc{RealizeAssignment}(S, f)$ outputs a set of triples $R$ instead of a valid DAG $G$. Then, by Lemma~\ref{lem:realize}, there is no DAG on $S$ that is consistent with the assignment of collider or non-collider given by $f$ on $R$. However, in line~\ref{line:create-ambiguities} of Algorithm~\ref{alg:orient}, all triples in $R$ must have been made ambiguous in $\Ehat$ and hence, the assignment $f$ on $R$ must be consistent with $\widetilde{G}^*$, which is a contradiction. Hence, $\textsc{RealizeAssignment}(S, f)$ must result in a valid DAG $\widehat{G}$ that realizes $f$ on $S$. Let $U$ be a vertex cover for the set of edges in $\widehat{G}$ between pairs of vertices that are not adjacent in $G^*$ and let $\widehat{G}'$ be the DAG obtained from deleting all edges incident to $U$. Crucially, note that deleting all edges incident to some set of vertices can never create a new unshielded triple, so every unshielded triple in $\widehat{G}'$ must also be an unshielded triple in $\widehat{G}$. Now, suppose that there exists $(\{i,j\},Z) \in \cA$ such that $(i \not \perp j \mid Z)_{\widehat{G}'}$. Let $W(\pi)$ be a witness graph for a path $\pi$ that d-connects $i,j$ given $Z$ in $\widehat{G}'$. Then, $\pi$ is also a d-connecting path in $\widehat{G}$, so all unshielded triples in $W(\pi)$ must have been made ambiguous in line~\ref{line:create-ambiguities-2} of Algorithm~\ref{alg:orient}. Therefore, they have the same collider/non-collider assignment as in $G^*$. But then, $(i \not \perp j \mid Z)_{G^*}$ by Lemma~\ref{lem:d-connection}, which is a contradiction to our assumptions. Hence, we must have $(i \perp j \mid Z)_{\widehat{G}'}$ for all $(i,j,Z) \in \cA$.
\end{proof}

Now, we just have to put all the pieces together:

\mainthm*

\begin{proof}[Proof]
    Let $\Ihat$ be the list of estimated conditional independencies after testing with Algorithm~\ref{alg:fdr} in line~\ref{CI-line} of Algorithm~\ref{alg:main}. We define $\Phat = \cP(G^*) \cap \Ihat$. By Lemma~\ref{lem:parentseparation}, the relations in $\Phat$ are entailed as d-separations by $G^*$. Let $\widehat{G}$ be the DAG represented by $\Ehat$ whose existence is guaranteed by Proposition~\ref{prop:CIstatements} with respect to $\cA = \Phat$. We want to bound the false positive error that $\widehat{G}$ has with respect to $G^*$ in terms of the number of false positive parent statements $\FPP(\Ihat, G^*) = |\cP(G^*) \setminus \Phat|$. First, consider the DAG $\widehat{G}'$ that is obtained from deleting all edges in $\widehat{G}$ that are incident to vertices of total degree at least $\Delta + 1$. Next, delete all edges incident to vertices $i$ such that $(i \not \perp j \mid \Pa(j, G^*))_{\widehat{G}'}$ and $(\{i,j\}, \Pa(j, G^*)) \notin \Phat$. In particular, if $\Ghat$ had a false positive adjacency $\{i,j\}$ with respect to $G^*$ with $i<j$, then we must have $(\{i,j\}, \Pa(j, G^*)) \notin \Phat$, so all edges incident to $i$ are deleted in $\widehat{G}''$. By Proposition~\ref{prop:CIstatements}, the resulting DAG $\widehat{G}''$ must now entail all true parent statements in $\cP(G^*)$ as d-separations, and hence $[\widehat{G}''] \preceq [G^*]$ by Lemma~\ref{lem:char_inclusion}. This implies that the false positive error of $\widehat{G}$ with respect to $G^*$ is upper-bounded by the number of edges that we deleted from $\widehat{G}$ to obtain $\widehat{G}''$. By assumption~\ref{ass:sparsity}, whenever a vertex $i$ has total degree $\Delta + k$ in $\widehat{G}$, it has at least $k$ false positive adjacencies with respect to $G^*$, so there are at least $k$ false positive parent statements in $\cP(G^*) \setminus \Phat$ involving $i$. Hence, we deleted at most $\Delta +1$ edges per such false positive parent statement to obtain $\widehat{G}'$ and $\Delta $ edges per remaining false positive parent statement to obtain $\widehat{G}''$. Hence, we get 
    \begin{align*}
        \FD(\Ehat, [G^*]) \leq \FD([\widehat{G}], [G^*]) \leq (\Delta + 1) \cdot \FPP(\Ihat, G^*).
    \end{align*}
    Note that according to how the adjacencies of $\Ehat$ are determined in lines~\ref{start-skeleton} and~\ref{end-skeleton}, we have $\Ad(\Ehat) = \Ad(\Ihat)$. Now, suppose the output $\widehat{H}$ of $\textsc{UndirectedGraphLearning}(\cS)$ contains $H^*$. Then, by Assumption~\ref{ass:mccf} and Proposition~\ref{prop:distance-m}, we know that the list $\cL$ of conditional independence tests that are conducted by Algorithm~\ref{alg:main} contains $\cP(G^*)$. Hence, by Assumption~\ref{ass:p-values} and Proposition~\ref{prop:fdr}, we get
    \begin{align*}
        \E\left[\FDR(\Ehat, [G^*]) \middle| \widehat{H} \supseteq H^*\right] &= \E\left[\frac{\FD(\Ehat, [G^*])}{\Ad(\Ehat)} \middle| \widehat{H} \supseteq H^*\right]\\
        &\leq \E\left[\frac{(\Delta+1) \cdot \FPP(\Ihat, G^*)}{\Ad(\Ihat)} \middle| \widehat{H} \supseteq H^*\right]\\
        &\leq (\Delta+1) \cdot \alpha
    \end{align*}
    Since the false discovery rate is bounded by $1$, we conclude using Assumption~\ref{ass:uglearning}:
    \begin{align*}
        \E\left[\FDR(\Ehat, [G^*])\right] \leq \E\left[\FDR(\Ehat, [G^*]) \middle| \widehat{H} \supseteq H^*\right] + \Pr\left(\widehat{H} \not\supseteq H^*\right) \leq (\Delta+1) \cdot \alpha + p.
    \end{align*}
\end{proof}

Finally, we prove the following short result that is mentioned in our complexity analysis of the $m$-conservative PC algorithm:

\begin{lemma} \label{lem:degree}
    Let $(A, \bv)$ be a linear Gaussian SEM with corresponding DAG $G$ and undirected graph $H$. If the total degree of $G$ is at most $\Delta$, then the total degree of $H$ is at most $\Delta^2$.
\end{lemma}

\begin{proof}
    By Lemma~\ref{lem:faithful_triangles}, we have that $\{i,j\}$ is an edge in $H$ if and only if 
    \begin{align*}
    \sum_{k \in C(i,j)} A_{ki}v_k^{-1}A_{kj} - A_{ji}v_j^{-1} \neq 0.
\end{align*}
Hence, we must have either $A_{ji} \neq 0$ or $A_{ki}A_{kj} \neq 0$ for some $k$, which means that $i,j$ have (undirected path) distance at most $2$ in $G$. Since each vertex $i$ has at most $\Delta$ neighbors in $G$, which each have at most $\Delta -1$ additional neighbors, we get that $i$ can be connected to at most $\Delta + \Delta(\Delta-1) = \Delta^2$ many vertices in $H$. 
\end{proof}

\section{Experiment details} \label{apx:experiments}
We use the following method for sampling random linear Gaussian SEMs on $[n]$ with expected degree $d$:
\begin{enumerate}
    \item sample a uniformly random ordering on $[n]$;
    \item sample a random acyclic graph $G$ by drawing each directed edge that is consistent with the ordering independently with probability $d / (n-1)$;
    \item for each edge $i \to j$ in $G$, sample the edge weights $A_{ji}$ independently from a uniform distribution on $[-2, -0.5] \cup [0.5, 2]$;
    \item for each $i$, sample the error variances $v_i$ independently from an exponential distribution with mean $1$;
    \item normalize edge weights and error variances so that each variable has unit variance.
\end{enumerate}

For the experiments on the graphs in Figures~\ref{fig:example1},~\ref{fig:example2}, and~\ref{fig:example3}, we use the same method of drawing edge weights for the free parameters, and then solve the equations described in the examples that introduce the unfaithful conditional independencies for the remaining parameters. For the Causal Chamber experiment, we used 11 variables from the observational dataset \texttt{lt\_crl\_benchmark\_v1/buchholz\_1\_obs} from the paper of~\citet{gamella_sanity_2025}. These 11 variables are five actuator settings that are randomly drawn according to a clipped linear Gaussian SEM and six sensor measurements in a controlled optical system. The relationships between the actuators and the sensors are partly non-linear. 

For the experiments in Figure~\ref{fig:experiments}, we use the $m$-conservative PC algorithm with $m=1, \alpha=0.1$ and the PC and CPC algorithm with significance level $\alpha=0.1$. For the Precision-Recall curves, we use adaptive grids, varying the significance level $\alpha$ for the constraint-based algorithms and the edge penalty $\lambda$ for the GES algorithm. We remark that we also had to use values of $\alpha$ larger than $1$ to get the $m$-conservative PC algorithm to output graphs of all densities up to the complete graph (this is a valid parameter setting, it just results in an empty theoretical guarantee). All constraint-based algorithms use the Fisher $z$-test for conditional independence testing and the $m$-conservative PC algorithm uses the neighborhood selection method for undirected graph learning by \citet{MB06} with Lasso penalty $0.2$.

For evaluation, we define precision as $1-\FDR$ and recall as $1-\FNR$. Analogous to $\FDR$, the false negative rate $\FNR(\widehat{\C}, \C^*)$ for an estimated CPDAG $\widehat{\C}$ with respect to a true CPDAG $\C^*$ is defined as 

\begin{align*}
    \FNR(\widehat{\C}, \C^*) :=  \frac{\Ad(\C^*)- \max_{\A \preceq \widehat{\C}, \A \preceq \C^*}\Ad(\A)}{\max(\Ad(\C^*), 1)}.  
\end{align*}

We remark that the numerator in both $\FDR$ and $\FNR$ is hard to evaluate, and has to be approximated heuristically for larger graphs in practice. However, our heuristic computation always produces a valid upper bound for false positive and false negative error, which translates into valid lower bounds for precision and recall. 

Finally, we report empirical runtimes of the $m$-conservative PC algorithm for datasets of 1000 samples each from randomly drawn linear Gaussian SEMs of varying sizes in Table~\ref{tab:mcpc-scalability}. All experiments were run on a MacBook Pro with a 10-core Apple M5 processor and 32 GB of memory.

\begin{table*}[t]
\centering
\caption{Runtimes of the $m$-conservative PC algorithm for random linear Gaussian SEMs with $n$ variables and expected degree $d$.}
\label{tab:mcpc-scalability}
\small
\setlength{\tabcolsep}{5pt}
\begin{tabular}{rrrrr}
\toprule
$n$ & $d$ & \shortstack{Median time (s)} & \shortstack{Completed fits}
 & \shortstack{Timeouts ($>300$s)} \\
\midrule
10 & 1 & 0.00788 & 10/10 & 0 \\
10 & 2 & 0.0269 & 10/10 & 0 \\
10 & 3 & 0.159 & 10/10 & 0 \\
10 & 4 & 1.47 & 10/10 & 0 \\
10 & 5 & 5.82 & 9/10 & 1 \\
\addlinespace[2pt]
12 & 1 & 0.0101 & 10/10 & 0 \\
12 & 2 & 0.0701 & 10/10 & 0 \\
12 & 3 & 1.15 & 9/10 & 1 \\
12 & 4 & 8.33 & 9/10 & 1\\
12 & 5 & 3.58 & 8/10 & 2\\
\addlinespace[2pt]
15 & 1 & 0.0165 & 10/10 & 0 \\
15 & 2 & 1.89 & 10/10 & 0 \\
15 & 3 & 7.94 & 8/10 & 2 \\
15 & 4 & 7.13 & 6/10 & 4 \\
15 & 5 & 155 & 1/10 & 9 \\
\addlinespace[2pt]
18 & 1 & 0.0298 & 10/10 & 0 \\
18 & 2 & 2.8 & 10/10 & 0 \\
18 & 3 & 14.1 & 8/10 & 2 \\
18 & 4 & 32.7 & 7/10 & 3 \\
18 & 5 & 71.5 & 1/10 & 9 \\
\addlinespace[2pt]
20 & 1 & 0.0605 & 10/10 & 0 \\
20 & 2 & 6.66 & 8/10 & 2 \\
20 & 3 & 69 & 5/10 & 5 \\
20 & 4 & 107 & 2/10 & 8 \\
20 & 5 & 9.25 & 1/10 & 9 \\
\addlinespace[2pt]
25 & 1 & 0.0579 & 9/10 & 1 \\
25 & 2 & 26.5 & 5/10 & 5 \\
25 & 3 & 61.6 & 1/10 & 9 \\
25 & 4 & 109 & 2/10 & 8 \\
25 & 5 & \textemdash & 0/10 & 10 \\
\bottomrule
\end{tabular}
\end{table*}

\end{document}